\documentclass[journal]{IEEEtran}

\usepackage{amsmath,amsfonts}
\usepackage{algorithmic}
\usepackage{algorithm}
\usepackage{array}
\usepackage[caption=false,font=normalsize,labelfont=sf,textfont=sf]{subfig}
\usepackage{textcomp}
\usepackage{stfloats}
\usepackage{url}
\usepackage{verbatim}
\usepackage{graphicx}
\usepackage{cite}
\usepackage{amssymb}
\usepackage{stmaryrd}
\usepackage{multirow}
\usepackage{graphics,color,psfrag}
\usepackage{booktabs}
\usepackage{algorithmic}
\usepackage{hyperref}

\newcommand{\bc}[1]{\mbox{\boldmath $\mathcal{#1}$}}

\newcommand{\mb}[1]{\mathbf{#1}}
\newcommand{\F}{\mathrm{F}}

\begin{document}

\title{When Bayesian Tensor Completion Meets Multioutput Gaussian Processes: Functional Universality and Rank Learning}

\author{
    Siyuan~Li, Shikai~Fang, 
    Lei~Cheng,~\IEEEmembership{Member,~IEEE,}
    Feng~Yin,~\IEEEmembership{Senior Member,~IEEE,} 
    Yik-Chung~Wu,~\IEEEmembership{Senior Member,~IEEE,}
    Peter~Gerstoft,~\IEEEmembership{Fellow,~IEEE,} and
    Sergios~Theodoridis,~\IEEEmembership{Life~Fellow,~IEEE}
    \thanks{
    The work of Lei Cheng was supported in part by the National Natural Science Foundation of China under Grant 62371418, in part by the Fundamental Research Funds for the Central Universities (226-2025-00168).
    The work of Feng Yin was supported in part by the Shenzhen Science and Technology Program under Grant JCYJ20220530143806016. 
    The work of Sergios Theodoridis was supported in part by the European Union under Horizon Europe (grant No. 101136568 - HERON). (Corresponding author: Lei Cheng)
    
    Siyuan Li, Shikai Fang and Lei Cheng are with the College of Information Science and Electronic Engineering, Zhejiang University, Hangzhou 310027, China (e-mails: \{lisiyuan, fsk, lei\_cheng\}@zju.edu.cn). Lei Cheng is also with Zhejiang Provincial Key Laboratory of Multi-Modal Communication Networks and Intelligent Information Processing.
    
    Feng Yin is with the School of Science \& Engineering, The Chinese University of Hong Kong, Shenzhen 518172, China (e-mail: yinfeng@cuhk.edu.cn).
    
    Yik-Chung Wu is with the Department of Electrical and Electronic Engineering, The University of Hong Kong, Hong Kong (email: ycwu@eee.hku.hk).
    
    Peter Gerstoft is with Technical University of Denmark, 2800 Lyngby, Denmark and NoiseLab, University of California, San Diego, La Jolla, CA 92093 USA (e-mail: gerstoft@ucsd.edu).
    
    Sergios Theodoridis is with the HERON - Center of Excellence in Robotics, Athena R.C, Athens, Greece (email: stheodor@di.uoa.gr).
    }
    \thanks{This paper has supplementary downloadable material available at http://ieeexplore.ieee.org., provided by the author.}
}



\maketitle

\begin{abstract}
Functional tensor decomposition can analyze multi-dimensional data with real-valued indices, paving the path for applications in machine learning and signal processing. 
A limitation of existing approaches is the assumption that the tensor rank---a critical parameter governing model complexity---is known.
However, determining the optimal rank is a non-deterministic polynomial-time hard (NP-hard) task and there is a limited understanding regarding the expressive power of functional low-rank tensor models for continuous signals.
We propose a rank-revealing functional Bayesian tensor completion (RR-FBTC) method. 
Modeling the latent functions through carefully designed multioutput Gaussian processes, RR-FBTC handles tensors with real-valued indices while enabling automatic tensor rank determination during the inference process.
We establish the universal approximation property of the model for continuous multi-dimensional signals, demonstrating its expressive power in a concise format.
To learn this model, we employ the variational inference framework and derive an efficient algorithm with closed-form updates. Experiments on both synthetic and real-world datasets demonstrate the effectiveness and superiority of the RR-FBTC over state-of-the-art approaches. 
The code is available at {\href{https://github.com/OceanSTARLab/RR-FBTC}{https://github.com/OceanSTARLab/RR-FBTC}}.
\end{abstract}

\begin{IEEEkeywords}
	tensor decomposition, Bayesian learning, Gaussian process, variational inference
\end{IEEEkeywords}

\section{Introduction}
\IEEEPARstart{T}{ensor} decompositions, in various forms such as CANDECOMP/PARAFAC (CP) \cite{bro1997parafac,kolda2009tensor}, Tucker \cite{tucker1966some}, and Tensor Train/Ring\cite{oseledets2011tensor, zhao2016tensor}, are widely used as concise yet expressive representations for a range of multi-dimensional data completion tasks, including images \cite{chen2022reweighted, liu2012tensor}, physical fields \cite{ li2023striking, shrestha2022deep}, and wireless channels \cite{de2021channel, zhang2024integrated, cheng2015subspace}. 
Among various tensor models, CP decomposition is the most fundamental and serves as the foundation for developing algorithms---both Bayesian and non-Bayesian---for other tensor decomposition formats \cite{zhao2015bayesian, cheng2022towards, sidiropoulos2017tensor, tong2023bayesian, xu2023tensor}. 
Despite its simplicity, CP decomposition possesses a desirable uniqueness property \cite{ten2002uniqueness} and enables the identification of latent factors under mild conditions, thereby facilitating the extraction of interpretable knowledge from tensor data.
CP decomposition can exactly represent any polynomial, which underscores its central role in tensor analysis~\cite{hitchcock1927expression,kolda2009tensor}.

Standard tensor decompositions \cite{kolda2009tensor, sidiropoulos2017tensor} face a fundamental limitation in many real-world applications: they are designed to handle data indexed by discrete integers. In contrast, many practical scenarios---such as geographic modeling---involve data indexed by continuous coordinates like latitude, longitude, and depth. 
As a result, one cannot directly apply these methods and needs preprocessing steps such as discretization or interpolation to align the data onto a fixed-resolution grid.   
Unfortunately, these steps may introduce artifacts and distort the inherent structure of the data, potentially compromising their physical nature and leading to degraded performance.

To address this limitation, recent studies have extended standard tensor models to continuous data representation, giving rise to functional tensor models \cite{fang2024functional, luo2023low, chertkov2023black, hashemi2017chebfun, kargas2021supervised, cho2024separable, schmidt2009function}. The central idea is to treat each element of the tensor as a sample drawn from a multivariate function, which is assumed to be decomposed into a set of latent factor functions, analogous to the latent factor matrices in standard tensor decomposition. The key challenge then becomes how to effectively represent these factor functions. Early approaches used predefined continuous functions, such as Chebyshev polynomials \cite{hashemi2017chebfun} or trigonometric functions \cite{kargas2021supervised}, to represent the latent factors. While these methods are straightforward and easy to implement, they lack flexibility due to their reliance on fixed functional forms, limiting their ability to capture complex data patterns.

Recent works have employed neural network (NNs) to represent the factor functions \cite{luo2023low, cho2024separable}. Given the universal approximation property of NN \cite{hornik1989multilayer}, these models are theoretically well-suited to represent continuous data with intricate details. However, NNs tend to overfit noise, lack interpretability, require extensive hyperparameter tuning, and cannot provide uncertainty quantification.

Another approach leverages Gaussian processes (GPs) to represent the latent factor functions within a Bayesian learning framework \cite{schmidt2009function, fang2024functional}. These approaches offer greater interpretability and provide a way to incorporate uncertainty, enhancing the models' robustness against noise. Moreover, like NN, GPs possess universal approximation properties \cite{micchelli2006universal, lee2017deep}.

Nevertheless, existing functional tensor approaches \cite{fang2024functional, luo2023low, chertkov2023black, hashemi2017chebfun, kargas2021supervised, cho2024separable, schmidt2009function}, either based on neural networks or GPs, require careful tuning of the model complexity to achieve optimal performance in data completion tasks. 
A main challenge lies in finding the balance between complexity and accuracy. With limited and noisy data samples, models with high complexity tend to overfit the noise, while models with low complexity tend to underfit the signal. 
In tensor models, the complexity is determined by the associated rank, such as the multilinear rank in Tucker decomposition \cite{tucker1966some} or the TT-rank in Tensor Train \cite{oseledets2011tensor}. In this paper, we focus on the most fundamental CP decomposition \cite{bro1997parafac, kolda2009tensor}, where the complexity is determined by the so-called tensor rank. 


Unfortunately, determining the optimal tensor rank for CP decomposition is a non-deterministic polynomial-time hard (NP-hard) task \cite{kolda2009tensor}. 
Many approaches are proposed for automatic rank determination in both Bayesian and non-Bayesian frameworks. In Bayesian learning, sparsity-promoting priors (automatic relevance determination (ARD) priors, Gaussian-Gamma priors, and global-local shrinkage priors) drive redundant columns in the factor matrices to zero, with the remaining columns forming the estimated tensor rank \cite{zhao2015bayesian, cheng2022towards}. Non-Bayesian approaches rely on techniques such as core consistency, information-theoretic criteria, cross-validation strategies, or nuclear-norm-based convex relaxations to select the rank~\cite{bro2003new, wax2003detection, recht2010guaranteed}. 
When data deviate from the assumed model, a trade-off must be made between penalizing model complexity and capturing the underlying data structure. For non-Bayesian approaches, this trade-off is often controlled through hyperparameters or thresholds that require careful tuning, a computationally intensive process. Despite the tuning-free advantage of Bayesian approaches, extending automatic rank determination from discrete factor matrices to continuous functions is little explored, not to mention the lack of theoretical characterization of its expressive power for the general case of continuous tensors.

This work proposes a functional (continuous) Bayesian CP decomposition with {\it automatic tensor rank learning}.  Specifically, we extend the single-output GP modeling of latent functions \cite{fang2024functional} to multioutput GP modeling\cite{chen2023multivariate, williams2006gaussian, dai2024graphical}, necessitating the specification of two covariance matrices (column covariance matrix and row covariance matrix). Ref \cite{dai2024graphical} considers MOGPs, but is unrelated to tensor modeling.
We model the column covariance matrix as a kernel one, with elements representing the correlations within each latent function, and the row covariance matrix as a diagonal matrix with each element {\it controlling the power of each latent function}. This modeling enables automatic rank determination during inference by incorporating sparsity-aware modeling and driving most latent functions toward zero during inference.

Within this general framework, a number of existing matrix and tensor decompositions, including recent works \cite{fang2024functional, song2023tensor}, are {\it special cases} of our proposed model. Furthermore, we {\it theoretically} prove that our model, despite the fact that it follows the simple CP decomposition, is guaranteed to approximate any multi-dimensional {\it continuous} signal. Also, we derive a variational inference algorithm \cite{bishop2006pattern, blei2017variational, theodoridis2024machine} for this new probabilistic model and show that all the update steps can be formulated in closed forms. The resulting algorithm is termed as Rank-Revealing Functional Bayesian Tensor Completion (RR-FBTC). Extensive experiments on both synthetic and real-world datasets validate the effectiveness and superiority of the proposed method over state-of-the-art approaches.

The remainder of the paper is organized as follows. Sec. II briefly reviews Bayesian CP modeling and multioutput GPs. Sec. III presents the proposed rank-revealing functional Bayesian tensor modeling and the related theoretical analysis.
Sec. IV derives an efficient algorithm based on the variational inference framework. Numerical results and discussions are reported in Sec. V, followed by a conclusion in Sec. VI. 

\section{Preliminaries}
This section provides the necessary background on Bayesian CP decomposition and multioutput Gaussian process, which will be useful for developing our method.

\subsection{Notation}

Bold lowercase and uppercase letters (e.g., $\mathbf{x}$ and $\mathbf{X}$) are used to denote the vectors and matrices. 
Uppercase bold calligraphic letters (e.g., $\bc{X}$) are used to denote tensors, while uppercase calligraphic letters (e.g., $\mathcal{X}$) represent sets or spaces.
Symbols $\circ, \otimes, \circledast, \odot$ denote the outer, Kronecker, Hadamard, and Khatri-Rao (column-wise Kronecker) products respectively.
The Khatri-Rao products of a set of vectors is denoted as $\bigodot_{k=1}^K \mb{x}_k= \mb{x}_1 \odot \cdots \odot \mb{x}_K$.
$\|\cdot\|_\F$ and $\|\cdot\|_{\infty}$ are the Frobenius norm and infinity norm.  $\langle\cdot,\cdot\rangle$ denotes the inner product. 
$\left \lceil {\cdot} \right \rceil$ is the expectation $\mathbb E [\cdot]$.
$\mb{X}^{T}$ is the transpose of $\mb{X}$.
$\mb{X}^{-1}$ is the inverse of $\mb{X}$.
$\mb{X}_{(k)}$ represents the mode-$k$ unfolding of $\bc{X}$.  
$\mathbb{R}$ is the field of real numbers.
The vectorization is $\operatorname{vec}(\cdot)$.
$|\mathcal{S}|$ is the cardinality of set $\mathcal{S}$.
$\operatorname{det}(\cdot)$ and $\operatorname{tr}(\cdot)$ are the determinant and the trace of a matrix, respectively.

\subsection{Sparsity-promoting priors for Bayesian CP Decomposition}
Consider a $K$-mode tensor $\bc{X} \in \mathbb{R}^{d_1\times \cdots \times d_K}$, where $d_k$ is the dimension of the $k$-th mode. 
In the CP decomposition, tensor $\bc{X}$ is decomposed as a sum of rank-1 components. A rank-1 component is defined as the outer product of $K$ vectors, one from each mode, i.e., $\mb{u}_r^1 \circ \mb{u}_r^2 \circ \cdots \circ \mb{u}_r^K$, where $\mb{u}_r^k\in \mathbb{R}^{d_k}$ is a vector corresponding to the $k$-th mode.
The CP decomposition is then given by:
\begin{align}
    \bc{X} =\sum_{r=1}^R \mb{u}_r^1 \circ \mb{u}_r^2 \circ \cdots \circ \mb{u}_r^K \triangleq \llbracket \mb{U}^1, \cdots, \mb{U}^K \rrbracket,
    \label{eq: cpd}
\end{align}
where $\mb{U}^k=[\mb{u}_1^k,\cdots, \mb{u}_R^k] \in \mathbb{R}^{d_k \times R}$ is the latent factor matrix of the $k$-th mode and $R$ is the tensor rank.

To automatically determine the rank, previous works have proposed imposing sparsity-promoting priors on the columns of the involved factor matrices \cite{zhao2015bayesian, ermis2014bayesian, cheng2022towards}. Specifically, they assume statistical independence among the respective columns and assign a Gaussian-Gamma pair prior to each column, i.e.,
\begin{align}
    p(\{\mb{U}^k\}_{k=1}^K|\{\gamma_r\}_{r=1}^R)  \nonumber  &= \prod_{r=1}^R \prod_{k=1}^K \mathcal{N}(\mb{u}_r^k|\mb{0}, \gamma_r^{-1} \mb{I}),  \nonumber \\
    p(\{\gamma_r\}_{r=1}^R|\{a_r, b_r\}_{r=1}^R) &= \prod_{r=1}^R \text{Gam}(\gamma_r | a_r, b_r),
    \label{eq: bcp}
\end{align}
where $\gamma_r^{-1}$ is the variance of the $r$-th columns $\{\mb{u}_r^k\}_{k=1}^K$ in the factor matrices. The hyper-parameters $\{a_r, b_r\}_{r=1}^R$ are usually set to small values (e.g., $10^{-3}$) to obtain non-informative priors.
This hierarchical construction leads to heavy-tailed marginal distributions for the columns $\{\mb{u}_r^k\}$ \cite{cheng2022towards}, thus achieving a sparse modeling and automatic rank determination during inference. 

Note that our focus here is on the prior modeling in Bayesian CP, while the likelihood function is not explicitly discussed since it depends on the specific observation model. 
In addition to the Gaussian-Gamma prior, other sparsity-promoting priors such as the horseshoe, Generalized Hyperbolic (GH) and shrinkage priors can also be employed~\cite{hinrich2020probabilistic, bhattacharya2015dirichlet, van2019shrinkage}. 
In this work, we focus on the Gaussian-Gamma prior because it is the most widely adopted.

\subsection{MOGP}
The multioutput GP (MOGP) extends the standard GP to handle vector-valued (or multioutput) functions \cite{chen2023multivariate, williams2006gaussian, dai2024graphical}.
Mathematically, suppose we are given a vector-valued function $\mb{f}(z)$ defined over a domain $\mathcal{Z} \in \mathbb{R}$, which follows a MOGP. The process is characterized by a mean function $\boldsymbol{\mu}(z): \mathcal{Z} \rightarrow \mathbb{R}^{1\times d}$, a kernel function (also known as column covariance function) $\varsigma(z, z'): \mathcal{Z} \times \mathcal{Z} \rightarrow \mathbb{R}$, and a positive semi-definite parameter matrix (called row covariance matrix) $\mb{\Omega} \in \mathbb{R}^{d \times d}$. 
Then, for any finite set of inputs $\{z_1, \dots, z_n\}$, the corresponding vector-valued outputs $\mathbf{F} = [\mb{f}(z_1)^T, \cdots, \mb{f}(z_n)^T]^T \in \mathbb{R}^{n\times d}$ follow a joint matrix-variate Gaussian distribution \cite[Eq. (1)]{chen2023multivariate}:
\begin{align}
    & p(\mb{F}|\mb{M},\mb{\Sigma},\mb{\Omega}) = \mathcal{MN}(\mb{M},\mb{\Sigma},\mb{\Omega})  \nonumber\\
    &= (2\pi)^{-\frac{dn}{2}}\operatorname{det}(\mb{\Sigma})^{-\frac{d}{2}}\operatorname{det}(\mb{\Omega})^{-\frac{n}{2}} \times   \nonumber \\
    &\exp\left[ -\frac{1}{2} \operatorname{tr}\left(\mb{\Omega}^{-1}(\mb{F}-\mb{M})^T\mb{\Sigma}^{-1}(\mb{F}-\mb{M}) \right) \right], 
\end{align}
where $\mb{M}\in \mathbb{R}^{n\times d}$ with $\mb{M}_{ij}=\boldsymbol{\mu}_j(z_i)$ and $\mb{\Sigma}\in \mathbb{R}^{n\times n}$ with $\mb{\Sigma}_{ij}=\varsigma(z_i, z_j)$.
The column covariance matrix $\mb{\Sigma}$ captures the correlations between $\mb{f}(z_i)$ and  $\mb{f}(z_j)$, $i,j=1,\cdots,n$, while the row covariance matrix $\mb{\Omega}$ represents the correlations between $[\mb{f}(z_i)]_{d_1}$ and $[\mb{f}(z_i)]_{d_2}$, $d_1,d_2=1,\cdots,d$, which is assumed to be independent of the input $z_i$.
For simplicity, we denote this process as $\mb{f} \sim \mathcal{MGP}(\boldsymbol{\mu}(\cdot), \varsigma (\cdot, \cdot), \mb{\Omega})$.

Given a set of $n$ noisy observations $\{z_i\in \mathbb{R}, \mb{y}_i\in \mathbb{R}^{1 \times d}\}_{i=1}^n $$ \triangleq \{\mb{z}=[z_1, \cdots,z_n]^T, \mb{Y}=[\mb{y}_1^T,\cdots,\mb{y}_n^T]^T\}$, we model the data as follows:
\begin{align}
    \mb{y}_i &= \mb{f}(z_i) + \boldsymbol{\eta}_i, \nonumber\\ 
    \mb{f} &\sim \mathcal{MGP}(\boldsymbol{\mu}(\cdot), \varsigma (\cdot, \cdot), \mb{\Omega}),
\end{align}
where $\boldsymbol{\eta}_i\sim\mathcal{N}(\mb{0},\sigma^2\mb{I})$ is Gaussian noise with zero mean and variance $\sigma^2$.
Let $\mb{F}_*=[\mb{f}_{*1}^T, \cdots, \mb{f}_{*m}^T]^T$ represent the latent variables at the test inputs $\mb{z}_*=[z_{*1}, \cdots, z_{*m}]^T$. Assuming $\boldsymbol{\mu}(\cdot)=\mb{0}$ as it is common in GPs, the joint distribution of $\mb{Y}$ and $\mb{F}_*$ follows a matrix-variate Gaussian distribution \cite{chen2023multivariate}
\begin{align}
    \left[\begin{array}{c}
        \mb{Y} \\ \mb{F}_*
    \end{array}\right] 
    \sim \mathcal{M N}\left(\boldsymbol{0},\left[\begin{array}{ll}
\mb{K}_{\mb{z}, \mb{z}} + \sigma^2\mb{I} & \mb{K}_{\mb{z}, \mb{z}_*} \\
\mb{K}_{\mb{z}, \mb{z}_*}^T & \mb{K}_{\mb{z}_*, \mb{z}_*}+ \sigma^2\mb{I},
\end{array}\right], \mb{\Omega}\right)
\end{align}
where $\mb{K}_{\mb{z}, \mb{z}}$ and $\mb{K}_{\mb{z}_*, \mb{z}_*}$ denote the covariance matrix evaluated on the training and test inputs, respectively; $\mb{K}_{\mb{z}, \mb{z}_*}$ denotes the cross covariance matrix with $[\mb{K}_{\mb{z}, \mb{z}_*}]_{ij} = \varsigma(z_{i}, z_{*j})$.
Due to the Gaussian assumptions and the property that the conditional of a Gaussian is a Gaussian, the posterior distribution of $\mb{F}_*$ is a matrix-variate Gaussian, given by \cite[Eq.~(5)]{chen2023multivariate}
\begin{align}
    p(\mb{F}_*|\mb{z}, \mb{Y}, \mb{z}_*) = \mathcal{MN}(\hat{\mb{M}}, \hat{\mb{\Sigma}}, \hat{\mb{\Omega}}),
\end{align}
where
\begin{flalign}
    \hat{\mb{M}} &= \mb{K}_{\mb{z}, \mb{z}_*}^T \left(\mb{K}_{\mb{z}, \mb{z}} + \sigma^2\mb{I} \right)^{-1} \mb{Y}, \nonumber \\ 
    \hat{\mb{\Sigma}} &=  \mb{K}_{\mb{z}_*, \mb{z}_*}+ \sigma^2\mb{I} - \mb{K}_{\mb{z}, \mb{z}_*}^T \left(\mb{K}_{\mb{z}, \mb{z}} + \sigma^2\mb{I} \right)^{-1} \mb{K}_{\mb{z}, \mb{z}_*}, \nonumber \\ 
    \hat{\mb{\Omega}} &= \mb{\Omega}.
\end{flalign}
We consider one-dimensional inputs $z_i$, but it can be extended to the multi-dimensional input cases.
The matrix-variate Gaussian distribution has been generalized to the tensor-variate Gaussian distribution, which allows modeling of multi-dimensional outputs~\cite{hoff2011separable}.

\section{Bayesian Modeling and Theoretical Analysis}
\subsection{Rank-revealing functional Bayesian tensor modeling}
Despite the successes of CP decomposition in various areas, its applicability is restricted to data with integer indices because the latent components $\{\mb{u}_r^k\}$ are discrete-index vectors, as shown in \eqref{eq: cpd}. To effectively represent data indexed by continuous values, a natural approach is to extend the ``underlying'' latent components from discrete-index vectors to continuous functions. Such a transformation yields a functional/continuous tensor that maps real-valued indices to their corresponding values.

\textbf{Functional CP decomposition:}
Consider a $K$-mode continuous tensor $\bc{X}$, where the value at a real-valued index $\mb{i}=[i_1,\cdots,i_K]^T$ is denoted as $x_{\mb{i}}$.
Inspired by the element-wise form of the CP decomposition in \eqref{eq: cpd}, $x_{\mb{i}}$ can be represented as
\begin{align}
	x_{\mb{i}} = \sum_{r=1}^R \prod_{k=1}^K u_r^k(i_k),
    \label{eq:7}
\end{align}
where $\{u_r^k(\cdot):\mathbb{R}\rightarrow \mathbb{R}\}$ is a set of one-dimensional functions and $R$ is the tensor rank.
In this formulation, the CP decomposition corresponds to the case where the columns of the factor matrices are the samples of the latent functions evaluated at integer grid points.
Alternatively, \eqref{eq:7} can be formulated as: 
\begin{align}
    x_{\mb{i}} = \left[ \bigodot_{k=1}^K \mb{U}^k(i_k) \right ]\mb{1}_R,
\end{align}
where $\mb{1}_R = [1,\cdots,1]^T \in \mathbb{R}^R$; $\mb{U}^k(\cdot): \mathbb{R}\rightarrow \mathbb{R}^{1\times R}$ is a vector-valued function that maps the continuous index $i_k$ of mode $k$ to a latent row vector of size $R$, i.e., $\mb{U}^k(i_k)=[u_1^k(i_k),\cdots,u_R^k(i_k)]$.

\textbf{Likelihood design:}
Given $N$ observed data points, denoted as $\mathcal{D} = \{(\mb{i}^n, y_n)\}_{n=1}^N$,\footnote{For example, $\mb{i}^n$ represents the (longitude, latitude, depth) of the $n$-th point, while $y_n$ corresponds to the temperature at the respective location.} we assume that the observations are contaminated by independent and identically distributed (i.i.d.) Gaussian noise. This leads to the following likelihood expression:
\begin{align}
	y_n = x_{\mb{i}^n} + w_n,
\end{align}
where $w_n\sim \mathcal{N}(0, \tau^{-1})$ is Gaussian noise with variance $\tau^{-1}$.
This gives the likelihood expression:
\begin{align}
	\label{eq: like1}
    p(\{y_n\}|&\{\mb{U}^k(\cdot)\}, \{\mb{i}^n\} ,\tau)=\prod_{n=1}^N p\left(y_n|\{\mb{U}^k(i_k^n)\}, \tau\right)
    \nonumber \\ 
    &= \prod_{n=1}^N \mathcal{N}\left(y_n\bigg|\left[ \bigodot_{k=1}^K \mb{U}^k(i^n_k) \right ]\mb{1}_R, \tau^{-1}\right),
\end{align}
where $i^n_k$ is the $k$-th term of $\mb{i}^n$.
We assume Gaussian noise, as common in the related literature. Other noise models can be adopted, with modifications to the likelihood function~\cite{ermis2014bayesian, wang2020conditional}.
Given the observations $\{y_n\}$, the goal is to infer the latent function values at the observed indices, $\{\{\mb{U}^k(i_k^n)\}_{k=1}^K\}_{n=1}^N$, along with the noise variance $\tau^{-1}$.

\begin{figure}[t]
	\centering
	\includegraphics[width=1\linewidth]{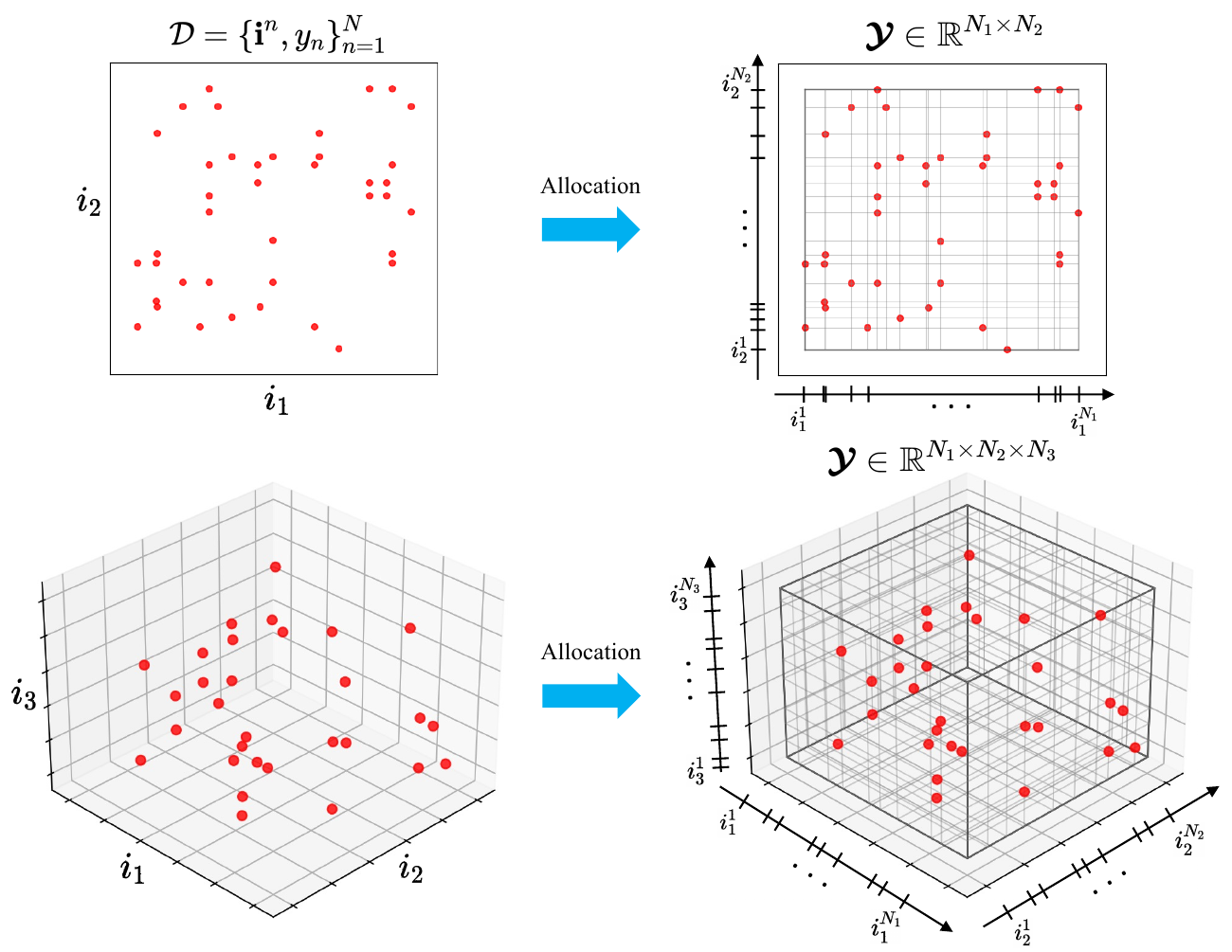}
	\caption{The allocation process in two- and three-dimensional scenarios, where $i_1$, $i_2$, and $i_3$ represent the axes of the modes (e.g., longitude, latitude, and depth). The observed data, indicated by red points, are allocated as entries of a discrete tensor $\bc{Y}$ (the gray box), in which the dimension of the $k$-th mode is $N_k=|\mathcal{S}_k|$, where $\mathcal{S}_k=\{i_k^1,\cdots,i_k^{N_k}\}$ is the real-valued coordinate set containing the unique values of mode-$k$'s indices from all the data.}
        \label{fig: reindex}
\end{figure}

To facilitate the inference of $\{\{\mb{U}^k(i_k^n)\}_{k=1}^K\}_{n=1}^N$, we construct a set of {\it ordered} real-valued coordinate sets $\{\mathcal{S}_k\}_{k=1}^K$, where each set $\mathcal{S}_k=\{i_k^1,\cdots,i_k^{N_k}\}$ contains the $N_k$ unique values of mode-$k$'s indices of all data in $\mathcal{D}$, sorted in ascending order, with $N_k=|\mathcal{S}_k|$.
Organizing the indices structured this way, the problem transforms into inferring the values of the latent functions at the new coordinate sets, $\{\mb{U}^k(\mathcal{S}_k)\}$. 
This property allows us to allocate the observations, $\{y_n\}$, to the entries of a discrete tensor $\bc{Y}\in \mathbb{R}^{N_1\times\cdots\times N_K}$.
Specifically, each observation $y_n$ is allocated to an entry $\bc{Y}_{n_1,\cdots,n_K}$, where 
\begin{align}
    n_k = \arg\min_j |i_k^n-i_k^j|, i_k^j\in \mathcal{S}_k, j=1,\cdots,N_k,
\end{align}
i.e., $i_k^n$ is the $n_k$-th smallest value in $\mathcal{S}_k$, $\forall i=1,\cdots,K$. This allocation process is presented in Fig.~\ref{fig: reindex}. Note that the grid $\mathcal{S}_k $ is formed so each point has a unique index. The irregular grids formed by the Cartesian product $\mathcal{S}_1\times \cdots \times \mathcal{S}_K$ (gray lines in Fig.~\ref{fig: reindex}) encompass all observed indices $\{\mb{i}^n\}$.


This allocation highlights the underlying tensor structure of the observations and enables a more compact likelihood, facilitating structured and efficient probabilistic inference.
Particularly, by assigning zero to the unobserved entries in $\bc{Y}$, the likelihood \eqref{eq: like1} is rewritten in a simpler form:
\begin{align}
    &p(\bc{Y}|\{\mb{U}^k(\mathcal{S}_k)\}_{k=1}^K,\tau) \nonumber\\ 
    &\propto \exp\!\!\left(\!-\frac{\tau}{2}\!\|\!\bc{O}\circledast(\bc{Y}-\llbracket \mb{U}^1(\mathcal{S}_1),\cdots,\mb{U}^K(\mathcal{S}_K) \rrbracket)\|_F^2\!\right),
\end{align}
where $\bc{O}\in \mathbb{R}^{N_1\times\cdots\times N_K}$ is an indicator tensor with $\bc{O}_{n_1,\cdots,n_K}=1$ if $\bc{Y}_{n_1,\cdots,n_K}$ is allocated with an observation data. 

\textbf{Rank-revealing functional prior\footnote{``Rank-revealing''\cite{chandrasekaran1994rank, li2005rank} indicates that the prior enables the model to automatically learn the tensor rank.}:} 
To achieve functional tensor modeling and automatic rank learning, we adopt the MOGP to model the latent functions, i.e., 
\begin{align}
	\mb{U}^k(\cdot) \sim \mathcal{MGP}(\mb{0}, \varsigma_k(\cdot,\cdot), \mb{\Gamma}^{-1}), \forall k=1, \cdots, K,
	\label{eq: mogp}
\end{align}
where $ \varsigma_k(\cdot,\cdot)$ is the kernel of mode $k$ and $\mb{\Gamma}^{-1}$ is a row variance matrix with $\mb{\Gamma}=\operatorname{diag}(\boldsymbol{\gamma})=\operatorname{diag}([\gamma_1,\cdots, \gamma_r])$.
Here, $\operatorname{diag}(\boldsymbol{\gamma})$ denotes a diagonal matrix with the elements of the vector $\boldsymbol{\gamma}$ along the main diagonal.
The interpretation of a diagonal structure of the row covariance matrix is that the latent basis functions $\{u_{r}^k(\cdot)\}_{r=1}^R$ are independent of each other, similar to the discrete-index cases \cite{zhao2015bayesian, cheng2022towards}.

We can specify the prior of the factor matrices $\{\mb{U}^k(\mathcal{S}_k)\}$.
Since these factor matrices are realizations of the MOGPs, the prior is\footnote{We omit ``$\forall k = 1, \dots, K$'' for simplicity. }
\begin{align}p(\mb{U}^k(\mathcal{S}_k)|\boldsymbol{\gamma})=\mathcal{MN}(\mb{0},\mb{\Sigma}_k,\mb{\Gamma}^{-1}),
	\label{eq: priorU}
\end{align}
where $\mb{\Sigma}_k \in \mathbb{R}^{N_k\times N_k}$ is the column covariance matrix with $[\mb{\Sigma}_k]_{i,j}=\varsigma_k(i,j)$.
For brevity, we omit the parentheses of $\mb{U}^k(\mathcal{S}_k)$ and use $\mb{U}^k$ in the following.

The input values (i.e., $i, j$) to the column covariance matrix, which is relevant to the continuity of the factor function, are {\it real-valued}, with integers as special cases.
Besides, the Gaussian prior has been widely used in previous Bayesian matrix/tensor modeling, e.g., \cite{zhao2015bayesian, cheng2022towards}.
As a result, many existing matrix/tensor decomposition methods are {\it special cases} of the proposed model, as elaborated in Sec.~\ref{secIII.B}.

\begin{figure}[t]
\centering
\includegraphics[width=0.99\linewidth]{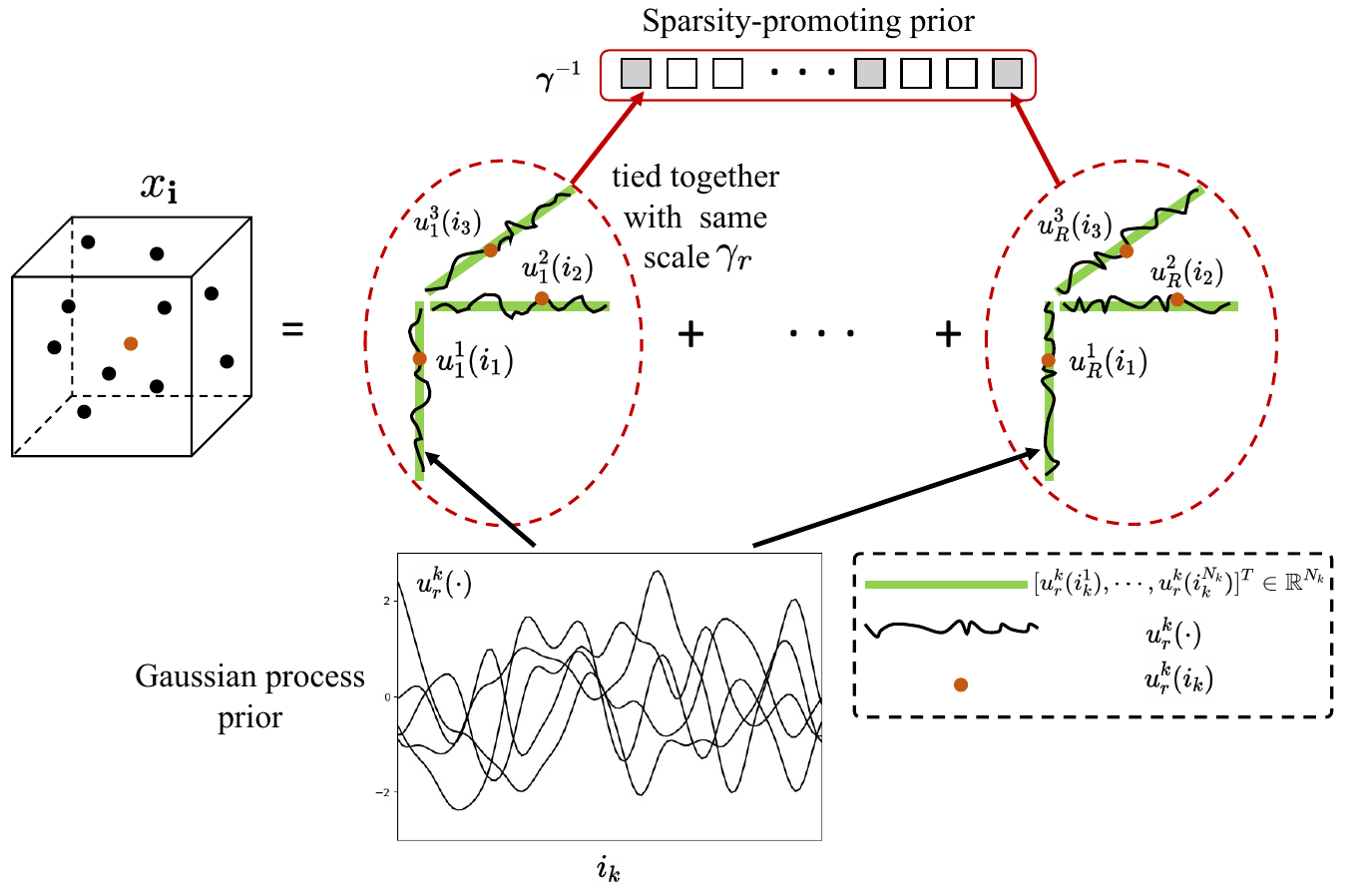}
\caption{The rank-revealing functional prior. The tensor entry $x_{\mb{i}}$, the brown point on the left side of the equal sign, is decomposed using the CP format \eqref{eq:7}, which is a sum of the products of latent functions $u_r^k(i_k)$. The black curves in red circles represent these latent functions $\{u^k_r(\cdot)\}$, which are modeled via MOGP priors \eqref{eq: mogp}. The green lines depict discrete vectors that contain the sampled function values at coordinate sets $\{\mathcal{S}_k\}$. The vector $\boldsymbol{\gamma}$ controls the powers of the processes across different modes, enabling rank-revealing functional modeling. An element of $\boldsymbol{\gamma}^{-1}$ near 0 (denoted by a blank box) results in the pruning of the corresponding rank-1 component during iteration. In contrast, a positive $\boldsymbol{\gamma}^{-1}$  signifies active components in the decomposition.
 }
        \label{fig: modeling}
\end{figure}


Since the row variance matrix has been assumed to be diagonal, the prior distribution \eqref{eq: priorU} simplifies to
\begin{align}
	p(\mb{U}^k|\boldsymbol{\gamma})&=\mathcal{MN}(\mb{0},\mb{\Sigma}_k,\mb{\Gamma}^{-1}) \nonumber\\ 
	&=\prod_{r=1}^R p(\mb{u}_r^k|\mb{\Sigma}_k,\gamma_r) =\prod_{r=1}^R \mathcal{N}(\mb{0},\gamma_r^{-1}\mb{\Sigma}_k),
\end{align}
where $\mb{u}_r^k \in \mathbb{R}^{N_k}$ is the $r$-th column of $\mb{U}^k$. 
This equation allows us to interpret the model as assigning one-dimensional GP priors to the latent functions $\{u^k_r(\cdot)\}$, with a kernel function (for column covariance matrix $\mb{\Sigma}_k)$ shared across $R$ rank-1 components. Additionally,  each diagonal element $\gamma_r$ controls the powers of the $r$-th stochastic processes $\{u_r^k(\cdot)\}$ at different modes. Consequently, the importance of the rank-1 components that contribute to the data tensor is determined by the vector $\boldsymbol{\gamma}$, enabling the rank-revealing property.
Inspired by Bayesian techniques in non-functional tensor models \cite{zhao2015bayesian, cheng2022towards}, we impose sparsity-promoting priors to the latent functions by assigning Gamma priors to the diagonal elements:
\begin{align}
    p(\boldsymbol{\gamma})&=\prod_{r=1}^R p(\gamma_r|a_r, b_r) = \prod_{r=1}^R\text{Gam}(\gamma_r | a_r, b_r),
\end{align}
where $\{ a_r, b_r\}_{r=1}^R$ are hyper-parameters.
We use the shape-rate formulation of the Gamma distribution.
For the noise variance parameter $\tau$, we assign it a conjugate Gamma prior:
\begin{align}
	p(\tau)=\text{Gam}(\tau|a_0,b_0),
\end{align}
where $a_0$ and $b_0$ are hyper-parameters.
An illustration of the proposed rank-revealing functional prior is in Fig.~\ref{fig: modeling}.
The red circles in Fig.~\ref{fig: modeling} represent a prior point decomposed in $R$ rank-1 modes. 

Let the parameter set be $\mb{\Theta}=\{\tau,\{\mb{U}^k\}, \{\gamma_r\}\}$. Then the joint distribution of the probabilistic model is
\begin{align}
	p(\mb{\Theta},\bc{Y})=p(\bc{Y}|\{\mb{U}^k\}_{k=1}^K,\tau)p(\tau)p(\boldsymbol{\gamma})\prod_{k=1}^K p(\mb{U}^k|\boldsymbol{\gamma}).
	\label{eq: jointpdf}
\end{align}

\subsection{Connection with other methods}
\label{secIII.B}
Many matrix and tensor decompositions, e.g., \cite{song2023tensor, fang2024functional, chen2023bayesian, zhao2015bayesian}, are special cases of the proposed RR-FBTC. Given that Gaussian noise assumptions are commonly made across the likelihood functions, our focus is primarily on the prior distributions of factor matrices. By examining these priors, we uncover the underlying connections and distinctions among these approaches. The connections between RR-FBTC and other existing matrix/tensor decomposition methods are summarized in Table \ref{tab:connection}.

First, the Correlated-CP method \cite{song2023tensor} models the intra-dimension correlations of the factor matrices using a set of covariance matrices. Its prior distribution is:
\begin{align}
    p(\mb{U}^k|\boldsymbol{\gamma})=\mathcal{MN}(\mb{0},\mb{C}_k,\mb{\Gamma}^{-1}),
\end{align}
where $\mb{C}_k$ is the column covariance matrix, capturing correlations within the columns of the factor matrix in mode $k$. The RR-FBTC reduces to the Correlated-CP when indices are discrete integers and the kernel matrix $\mb{\Sigma}_k$ is $\mb{C}_k$. Compared to this approach, RR-FBTC is more general as it accommodates real-valued indices. 

Second, the continuous tensor decomposition methods, FunBat \cite{fang2024functional} and LRTFR \cite{luo2023low}, are closely related to RR-FBTC. FunBat also uses GPs to represent continuous latent functions but does not include a rank learning mechanism. The proposed RR-FBTC reduces to FunBat if the row covariance matrix $\mb \Gamma^{-1}$ is an identity matrix. 
RR-FBTC extends the single-output GP modeling (in FunBat \cite{fang2024functional}) to the MOGP. This alters both the modeling and the inference procedure. By employing MOGPs, we jointly model the entire factor matrices using matrix-variate distributions, which enables automatic rank learning during inference, a capability absent in prior works and important for robustness under sparse and noisy conditions. The richer dependency structure requires a different inference strategy. We develop a closed-form variational inference algorithm tailored to MOGP-based continuous tensor modeling.
In LRTFR, neural networks, specifically multi-layer perceptrons (MLPs), represent the continuous latent functions. It is known that an MLP with i.i.d. random weights is equivalent to a GP, in the limit of infinite network width \cite{Neal1996}. Therefore, the LRTFR can be seen as a specific variant of RR-FBTC, with the kernel determined by the network. 

Third, BMCG \cite{chen2023bayesian}, a graph-guided matrix decomposition approach, models correlations in the factor matrix via a graph Laplacian matrix. The prior distribution in BMCG is:
\begin{align}
    p(\mb{U}^k|\boldsymbol{\gamma})=\prod_{r=1}^R \mathcal{N}(\mb{0},\gamma_r^{-1}\mb{\mb{L}}_k),
\end{align}
where $\mb{L}_k$ is the graph Laplacian matrix for the $k$-th mode. The RR-FBTC extends BMCG to handle multi-dimensional data with real-valued indices, using kernels to model the continuity of the latent factors instead of the Laplacian matrix.

Finally, from \eqref{eq: priorU} and \eqref{eq: bcp}, we observe that RR-FBTC reduces to the standard Bayesian matrix/tensor decomposition, where a sparsity-promoting prior distribution is placed on the factor matrices. In particular, the model reduces to the case where the columns of the factor matrices follow a Gaussian-Gamma prior when the column covariance matrix is reduced to the identity matrix and the indices are integers. However, these standard methods are designed for discrete settings and are not directly applicable to continuous tensors.

\begin{table*}[ht]
\caption{A summary of various matrix/tensor decomposition methods and their connections to the proposed approach.}
    \centering
    \begin{tabular}{p{2.2cm}p{2.2cm}p{2.2cm}p{2.0cm}p{2.2cm}p{2.2cm}p{2.3cm}}
        \toprule
        \textbf{model} & Correlated-CP\cite{song2023tensor} & FunBaT\cite{fang2024functional} & LRTFR\cite{luo2023low} & BMCG\cite{chen2023bayesian} & Bayesian CP\cite{zhao2015bayesian} & RR-FBTC\\
        \midrule
        discrete/continuous & discrete & continuous & continuous & discrete & discrete & continuous\\
        \midrule
        prior for $\mb{U}^k$ & $\mathcal{MN}(\mb{0},\mb{C}_k,\mb{\Gamma}^{-1})$ & $\mathcal{MN}(\mb{0},\mb{\Sigma}_k,\mb{I})$ & --- & $\mathcal{MN}(\mb{0},\mb{L}_k,\mb{\Gamma}^{-1})$ & $\mathcal{MN}(\mb{0},\mb{I},\mb{\Gamma}^{-1})$ & $\mathcal{MGP}(\mb{0}, \varsigma_k, \mb{\Gamma}^{-1})$\\
        \midrule
        connection to RR-FBTC & $\mb{\Sigma}_k=\mb{C}_k$, discrete & $\mb{\Gamma}=\mb{I}$, pre-defined tensor rank & deterministic variant   & $\mb{\Sigma}_k=\mb{L}_k$, 2D discrete & $\mb{\Sigma}_k=\mb{I}$, discrete & ---\\
        \bottomrule
    \end{tabular}
    \label{tab:connection}
\end{table*}

These connections demonstrate that RR-FBTC not only unifies existing methods but it also offers enhanced flexibility and generality, particularly for multi-dimensional and real-valued indices.
Adopting a more generalized modeling framework, we anticipate achieving greater adaptability and improved performance across diverse application scenarios, as demonstrated in Sec.~\ref{secV}.

\subsection{Theoretical analysis}
The RR-FBTC is based on the CP decomposition, thereby inheriting its related advantages, including mathematical simplicity and robustness to noise.
However, a pertinent question arises regarding whether this simplicity may compromise the expressiveness for continuous functions. 
To address this concern, we demonstrate that RR-FBTC has a universal approximation property (UAP), endowing it with high expressive power. 
To establish this, we first introduce some key definitions and lemmas.

\noindent \textbf{Definition 1} (RKHS, e.g., \cite{aronszajn1950theory}): \emph{Let $\mathcal{Z}$ be a nonempty set and $\varsigma$ be a positive definite kernel on $\mathcal{Z}$. A Hilbert space $\mathcal{H}$ of functions on $\mathcal{Z}$ equipped with an inner-product $\langle \cdot, \cdot \rangle_{H}$ is called a reproducing kernel Hilbert space (RKHS) with reproducing kernel $\varsigma$, if the following are satisfied:
\begin{itemize}
	\item For all $\mb{z}\in \mathcal{Z}$, we have $\varsigma(\cdot, \mb{z})\in \mathcal{H}$;
	\item For all $\mb{z}\in \mathcal{Z}$ and for all $f\in \mathcal{H}$,
	\begin{equation*}
		f(\mb{z})=\langle f, \varsigma(\cdot,\mb{z}) \rangle_{\mathcal{H}}~~~ (\text{Reproducing property}).
	\end{equation*}
\end{itemize}
}

It is well known that for every positive definite kernel $\varsigma$, there exists a unique RKHS $\mathcal{H}$ for which $\varsigma$ is the reproducing kernel \cite{steinwart2008support}.  Conversely, given a kernel, the corresponding RKHS is unique (up to isometric isomorphisms), meaning each kernel generates a distinct RKHS.

To generate an RKHS for a given kernel, consider the kernel function $\varsigma(\cdot,\cdot)$, a function of two variables. Suppose, for $N$ data points $\{\mb{z}_n\}_{n=1}^N$, we fix one variable to yield $\varsigma(\mb{z}_1,\cdot), \cdots, \varsigma(\mb{z}_N,\cdot)$.
Let $\mathcal{H}_0$ be a Hilbert space containing all possible linear combinations of these functions:
\begin{align}
	\mathcal{H}_0=\bigg\{ f(\cdot) = \sum_{i=1}^N c_i \varsigma(\mb{z}_i,\cdot): ~&c_i \in \mathbb{R}, \mb{z}_i \in \mathcal{Z}, \nonumber\\ 
    &\text{for}~ i=1,\cdots,N \bigg\}.
\end{align}
The inner-product of $\mathcal{H}_0$ is defined as follows: for any $f=\sum_{i=1}^na_i\varsigma(\cdot,\mb{z}_i) \in \mathcal{H}_0$ and $g=\sum_{j=1}^m b_j\varsigma(\cdot,\mb{z}_j)\in \mathcal{H}_0$, the inner-product is given by
\begin{align}
	\langle f,g \rangle_{\mathcal{H}_0}= \sum_{i=1}^n\sum_{j=1}^m a_ib_j\varsigma(\mb{z}_i,\mb{z}_j).
\end{align}
Then, the RKHS $\mathcal{H}$ corresponding to kernel $\varsigma$ is the closure of $\mathcal{H}_0$ with respect to the norm induced by the inner-product $\langle \cdot, \cdot\rangle_{\mathcal{H}_0}$ \cite{kanagawa2018gaussian}. This reveals that any function in the RKHS can be approximated with a linear combination of the kernel function.

\noindent \textbf{Definition 2} (Universal kernel, e.g., \cite{micchelli2006universal}): \emph{
Let $C(\mathcal{Z})$ denote the space of all continuous functions on an input space $\mathcal{Z}$. A continuous kernel $\varsigma$ on a compact metric space $\mathcal{Z}$ is called universal if the RKHS $\mathcal{H}$, with kernel function $\varsigma$, is dense in $C(\mathcal{Z})$. In other words, for every function $g \in C(\mathcal{Z})$ and all $\epsilon > 0$, there exists a function $f \in \mathcal{H}$ such that $\|f-g\|_{\infty} < \epsilon$.
}

We focus on the scenario where the space $\mathcal{Z}$ is a compact metric space. Given that every function in the RKHS can be approximated with a linear combination of kernel functions, we have the following lemma:

\noindent \textbf{Lemma 1:} \emph{Let $\varsigma$ be a universal kernel on a compact metric space $\mathcal{Z}$. For every function $g \in C(\mathcal{Z})$ and $\epsilon>0$, there exist coefficients $\{c_i\}_{i=1}^N$ and data points $\{\mb{z}_i\}_{i=1}^N$ such that 
\begin{align}
	\left\|\sum_{i=1}^N c_i \varsigma(\cdot,\mb{z}_i) -g\right\|_{\infty} < \epsilon.
\end{align}
}

\noindent \emph{Proof:} See App. A.

Lemma 1 follows directly from the definitions of universal kernels and RKHS, serving as an important component in proving the UAP of RR-FBTC.

\noindent \textbf{Remark 1} (Examples of universal kernels \cite{steinwart2008support, sriperumbudur2011universality}): \emph{
Let $\mathcal{Z}$ be a compact subset of $\mathbb{R}^D, \alpha > 0$, and $h > 0$. Then the following kernels on $\mathcal{Z}$ are universal:
\begin{itemize}
	\item exponential kernel: $\varsigma(\mb{z},\mb{z}')=\exp(\langle \mb{z},\mb{z}' \rangle)$,
        \item RBF kernel: $\varsigma(\mb{z},\mb{z}'):=exp(-h^{-2}\|\mb{z}-\mb{z}'\|_2^2)$
	\item Matern kernel: \\$\varsigma(\mb{z},\mb{z}')=\frac{1}{2^{\alpha-1} F(\alpha)}\left(\frac{\sqrt{2 \alpha}\left\|\mb{z}-\mb{z}^{\prime}\right\|}{h}\right)^\alpha K_\alpha\left(\frac{\sqrt{2 \alpha}\left\|\mb{z}-\mb{z}^{\prime}\right\|}{h}\right)$,
\end{itemize}
where $F$ is the gamma function, and $K_\alpha$ is the modified Bessel function of the second kind of order $\alpha$.
}

Kernel functions are closely related to GPs. In particular, the posterior mean function of a GP regression resides within the RKHS associated with the corresponding kernel \cite{kanagawa2018gaussian}. This implies that a GP equipped with a universal kernel inherits the UAP. 

\noindent \textbf{Definition 3} (Tensor product kernel, e.g., \cite{szabo2018characteristic}): \emph{A kernel $\varsigma$ is called a tensor product kernel if for $\mb{z},\mb{z}'\in \mathbb{R}^D$, it satisfies
\begin{align}
	\varsigma(\mb{z},\mb{z}')=\prod_{d=1}^D \varsigma(z_d,z'_d),
\end{align}
where $\mb{z}=[z_1,\cdots,z_D]^T$.
}

From the definition, it is clear that both the exponential and RBF kernels are tensor product kernels.
Now we present the theorem about the UAP of the proposed model.

\noindent \textbf{Theorem 1 (UAP of RR-FBTC):} \emph{ Let $\mathcal{Z}$ be a compact metric space in $\mathbb{R}^D$, and let $\varsigma$ be a universal tensor product kernel with $\varsigma(\mb{z},\mb{z}')=\prod_{d=1}^D \varsigma(z_d,z'_d)$. For any continuous function $g \in C(\mathcal{Z})$ and $\epsilon>0$, there exist data points such that the overall posterior mean function
\begin{align}
	f=\llbracket \bar{\mb{U}}^1,\cdots,\bar{\mb{U}}^D \rrbracket,
\end{align}
 satisfies $\left\|f -g\right\|_{\infty} < \epsilon$ provided the rank $R$ is sufficiently large.
Here, each latent function $u_r^d(\cdot)$ follows a GP prior $\mathcal{GP}(0, \varsigma)$
and $\bar{\mb{U}}^d=[\bar{u}_1^d, \cdots, \bar{u}_R^d]$ represents the posterior mean function for mode $d$.
}

\noindent \emph{Proof:} See App. A.

The UAP of RR-FBTC indicates that, despite its mathematical succinctness, it possesses the capability to approximate any continuous multi-dimensional function with arbitrary accuracy. 
The conciseness and expressiveness allow the model to effectively capture complex data patterns while also avoiding noise overfitting, which lays the foundation for its superior performance in various applications.

\section{Algorithm Development}
Given the probabilistic model in \eqref{eq: jointpdf}, the next step is estimating the posterior distribution $p(\boldsymbol{\Theta}|\bc{Y})$ of the latent parameters from the observation tensor $\bc{Y}$:
\begin{align}
	p(\boldsymbol{\Theta}|\bc{Y})=\frac{p(\bc{Y}|\boldsymbol{\Theta})p(\boldsymbol{\Theta})}{p(\bc{Y})}=\frac{p(\bc{Y}|\boldsymbol{\Theta})p(\boldsymbol{\Theta})}{\int p(\bc{Y}|\boldsymbol{\Theta})p(\boldsymbol{\Theta}) d\boldsymbol{\Theta}},
\end{align}
where $\mb{\Theta}=\{\tau,\{\mb{U}^k\}, \{\gamma_r\}\}$.	
Computing the posterior distribution $p(\boldsymbol{\Theta}|\bc{Y})$ requires integrating over the parameter space of $\boldsymbol{\Theta}$. However, exact evaluation is analytically intractable due to the the high dimensionality of the latent space and the nonlinear coupling among the parameters. 
Consequently, we resort to approximate Bayesian inference methods. 
We employ variational inference (VI) for its computational efficiency and suitability for high-dimensional inference problems, and it is guaranteed to converge to a local optimum\cite{bishop2006pattern, blei2017variational, theodoridis2024machine}.

VI approximates the true posterior distribution $p(\boldsymbol{\Theta}|\bc{Y})$ with a variational distribution $q(\boldsymbol{\Theta}) \in \mathcal{F}$ by minimizing the Kullback-Leibler (KL) divergence:
\begin{align}
    \mathrm{KL}(q(\mathbf{\Theta}) \| p(\mathbf{\Theta} | \bc{Y})) &=\int q(\mathbf{\Theta}) \ln 
    \left(\frac{q(\mathbf{\Theta})}{p(\mathbf{\Theta} | \bc{Y})}\right) d \mathbf{\Theta} \nonumber \\
    &=\ln p(\bc{Y})-\mathcal{L},
\end{align}
where $\ln p(\bc{Y})$ is the evidence and $\mathcal{L}$ is the evidence lower bound (ELBO) defined as 
$\mathcal{L}=\int q(\mathbf{\Theta}) \ln \left(\frac{p(\bc{Y}, \mathbf{\Theta})}{q(\mathbf{\Theta})}\right) d \mathbf{\Theta}$. 
Since the evidence is constant, minimizing the KL divergence is equivalent to maximizing the ELBO. 
Consider $\mathcal{F}$ to be the mean-field family, i.e., $q(\mathbf{\Theta})=\prod_{j} q\left(\mathbf{\Theta}_{j}\right)$, 
where $\mathbf{\Theta}_{j}\subset \mathbf{\Theta}$ with $\cup\mathbf{\Theta}_{j}=\mathbf{\Theta} $ and 
$\cap\mathbf{\Theta}_{j}=\varnothing $. Then the optimal variational probability density function (pdf) for $\mathbf{\Theta}_{j}$ is given by
\begin{align}
    \ln q\left(\mathbf{\Theta}_{j}\right)=
    \left \lceil \ln p(\bc{Y}, \mathbf{\Theta})\right \rceil_{\mathbf{\Theta} \backslash \mathbf{\Theta}_{j}}+\text{ const},
    \label{var:pdf}
\end{align}
where $\mathbf{\Theta} \backslash \mathbf{\Theta}_{j}$ denotes the set 
$\mathbf{\Theta}$ with $\mathbf{\Theta}_{j}$ removed.


Since the computation of the variational distribution of one variable group $\mathbf{\Theta}_{j}$ in \eqref{var:pdf} depends on the distributions of the other variables $\mathbf{\Theta} \backslash \mathbf{\Theta}_{j}$, the variational distributions is updated iteratively.
We derive closed-form updates for each variable group under the mean-field assumption $q(\mathbf{\Theta})=q(\boldsymbol{\gamma})q(\tau)\prod_{k=1}^K \prod_{r=1}^R q(\mb{u}_r^k)$.
Only the final results are presented, the derivations are in App. B.

The optimal variational pdf $q(\mb{u}_r^k)$ is a normal distribution
\begin{align}
    q(\mb{u}_r^k)=\mathcal{N}(\mb{u}_r^k|\mb{m}_r^k,\boldsymbol{\Psi}_r^k),
\end{align}
with the mean and variance given by (see (B.6))
\begin{align}
	\mb{m}_r^k &= \lceil\tau\rceil \boldsymbol{\Psi}_r^k \bigg[\bc{O}\circledast(\bc{Y}-\sum_{\substack{s=1\\ s\neq r}}^R\lceil\mb{u}_s^1\circ\cdots\circ \mb{u}_s^K \rceil)\bigg]_{(k)}\bigodot_{\substack{l=K\\ l\neq k}}^1 \lceil\mb{u}_r^l\rceil,\\
	\boldsymbol{\Psi}_r^k &= \bigg\{ \lceil \tau \rceil \operatorname{diag}\bigg[\bc{O}_{(k)} \bigodot_{\substack{l=K\\ l\neq k}}^1 \lceil\mb{u}_r^l \circledast \mb{u}_r^l\rceil \bigg] + \lceil\gamma_r\rceil\mb{\Sigma}_k^{-1} \bigg\}^{-1},
    \label{vi:phi}
\end{align}
where $\bigodot_{\substack{l=K\\ l\neq k}}^1 \lceil\mb{u}_r^l\rceil=\lceil\mb{u}_r^K\rceil \odot \cdots \odot\lceil\mb{u}_r^{k+1}\rceil \odot \lceil\mb{u}_r^{k-1}\rceil \odot \cdots \\ \lceil\mb{u}_r^{1}\rceil$.

The optimal variational pdf for $\boldsymbol{\gamma}$ is (see (B.9))
\begin{align}
	q(\boldsymbol{\gamma})=\prod_{r=1}^R \text{Gam}(\gamma_r|\hat{a}_r,\hat{b}_r),
\end{align}
with
\begin{align}
	\hat{a}_r = a_r+\frac{1}{2}\sum_{k=1}^K N_k,\quad
	\hat{b}_r = b_r + \frac{1}{2}\sum_{k=1}^K\lceil (\mb{u}_r^k)^T\mb{\Sigma}_k^{-1}\mb{u}_r^k \rceil.
\end{align}

The optimal variational pdf $q(\tau)$ is (see (B.12))
\begin{align}
	q(\tau) = \text{Gam}(\tau|\hat{a}_0,\hat{b}_0),
\end{align}
with
\begin{align}
	\hat{a}_0 \!=\! a_0 + \frac{N}{2},\quad
	\hat{b}_0 \!=\! b_0 + \frac{1}{2} \lceil \left\| \bc{O}\circledast(\bc{Y}-\llbracket \mb{U}^1,\cdots,\mb{U}^K \rrbracket) \right\|_F^2 \rceil.
\end{align}
The procedure is summarized in \textbf{Algorithm 1}. The computational complexity of the proposed algorithm is dominated by the matrix inverse in \eqref{vi:phi}. Thus the overall complexity is $\mathcal{O}(R_{\text{init}}\max(N_1^3, \cdots, N_K^3))$ where $R_{\text{init}}$ is the initial rank.
The estimated rank rapidly decreases to a small value within a few iterations, and the model typically converges quickly. Moreover, acceleration techniques such as Nystr\"om or random Fourier feature approximations can be employed to approximate the kernel matrix~\cite{eriksson2018scaling, liu2020gaussian, yin2020linear, jazbec2021scalable}. Runtime comparisons are in App. C.

\begin{algorithm}[t]
    \caption{Rank-revealing functional Bayesian tensor completion (RR-FBTC).}
    \label{alg:RR-FBTC}
    \renewcommand{\algorithmicrequire}{\textbf{Input:}}
    \renewcommand{\algorithmicensure}{\textbf{Output:}}
    \begin{algorithmic}[1]
        \REQUIRE Dataset $\mathcal{D}=\{(\mb{i}^n,y_n)\}_{n=1}^N$, where $\{\mb{i}^n\}$ are real-valued indices; kernel parameters.  
        \STATE  Build the real-valued coordinate sets $\{\mathcal{S}_k\}_{k=1}^K$ and allocate the data to the entries of tensor $\bc{Y}$.
        \STATE  Build the indicator tensor $\bc{O}$.
        \STATE  Initialize tensor rank $R_{\text{init}}$ and parameters in the variational distributions.
        \WHILE{not converge}
            \STATE Update $q(\mb{u}^k_r)$ via (30)-(32) for $k=1,\cdots,K$ and $r=1,\cdots,R$.
            \STATE Update $q(\boldsymbol{\gamma})$ via (33)-(34) for $k=1,\cdots,K$.
            \STATE Update $q(\tau)$ via (35)-(36).
        \ENDWHILE
    \ENSURE Learned variational distribution $q(\mathbf{\Theta})=q(\boldsymbol{\gamma})q(\tau)\prod_{k=1}^K \prod_{r=1}^R q(\mb{u}^k_r)$, based on which any unobserved entry can be derived via (37)-(39).    
    \end{algorithmic}
\end{algorithm}

After completing the inference process, we derive the value of the continuous tensor at any unseen index.
Given a set of test indices $\{\underline{\mb{i}}^1, \cdots, \underline{\mb{i}}^M\}$,  let the test indices for mode $k$ be denoted as $\underline{\mb{i}}_k=[\underline{i}^1_k,\cdots,\underline{i}^M_k]^T$, with the corresponding variable represented by $\underline{\mb{U}}^k=[\mb{U}^k(\underline{i}^1_k);\cdots;\mb{U}^k(\underline{i}^M_k)]$.  The posterior pdf of $\underline{\mb{U}}^k$ is then derived using results from MOGP \cite{chen2023multivariate, williams2006gaussian, dai2024graphical}:
\begin{align}
    p(\underline{\mb{U}}^k|\mb{i}_k, \mb{M}^{(k)}, \underline{\mb{i}}_k) = \mathcal{MN}(\hat{\mb{M}}_k, \hat{\mb{\Sigma}}_k, \hat{\mb{\Omega}}_k),
\end{align}
where
\begin{flalign}
    \hat{\mb{M}}_k &= \mb{K}_{\mb{i}_k, \underline{\mb{i}}_k}^T \left(\mb{K}_{\mb{i}_k, \mb{i}_k} + \sigma^2\mb{I} \right)^{-1} \mb{M}^{(k)}, \nonumber \\ 
    \hat{\mb{\Sigma}}_k &=  \mb{K}_{\underline{\mb{i}}_k, \underline{\mb{i}}_k}+ \sigma^2\mb{I} - \mb{K}_{\mb{i}_k, \underline{\mb{i}}_k}^T \left(\mb{K}_{\mb{i}_k, \mb{i}_k} + \sigma^2\mb{I} \right)^{-1} \mb{K}_{\mb{i}_k, \underline{\mb{i}}_k}, \nonumber \\ 
    \hat{\mb{\Omega}}_k &= \mb{\Lambda}^{-1}.
\end{flalign} 
Here, $\mb{i}_k=[i_k^1,\cdots,i_k^{N_k}]^T \in \mathbb{R}^{N_k}$ represents the observed index vector for mode $k$; $\mb{K}_{\mb{i}_k, \mb{i}_k} \in \mathbb{R}^{N_k\times N_k}$ and $\mb{K}_{\underline{\mb{i}}_k, \underline{\mb{i}}_k} \in \mathbb{R}^{M\times M}$ are the corresponding covariance matrices; $\mb{K}_{\mb{i}_k, \underline{\mb{i}}_k} \in \mathbb{R}^{N_k\times M}$ is the cross covariance matrix; $\mb{M}^{(k)}=[\mb{m}_1^k, \cdots, \mb{m}_R^k] \in \mathbb{R}^{N_k\times R}$ is the posterior mean matrix; and $\boldsymbol{\Lambda}\in \mathbb{R}^{R\times R}$ is a diagonal matrix with $\boldsymbol{\Lambda}_{rr}=\lceil \gamma_r \rceil$.

Finally, the estimated value of the continuous tensor at a new test index $\underline{\mb{i}}^m$ is obtained as:
\begin{align}
    x_{\underline{\mb{i}}^m} = \left[ \bigodot_{k=1}^K \hat{\mb{M}}_k(\underline{i}_k^m) \right ]\mb{1}_R, \forall m=1,\cdots, M,
\end{align}
where $ \hat{\mb{M}}_k(\underline{i}_k^m)$ is the $\underline{i}_k^m$-th row of $\hat{\mb{M}}_k$.

\noindent \textbf{Remark 2}: The column covariance $\hat{\mb{\Sigma}}_k$ captures the uncertainty in the latent function at the test indices, reflecting the confidence of the predictions. The row covariance $\hat{\mb{\Omega}}_k$, as a diagonal matrix, controls the scale of the latent function, determining the power of each rank-1 component. Unlike neural network-based tensor decomposition \cite{luo2023low, cho2024separable} that yield only point estimates, RR-FBTC provides a probabilistic characterization with uncertainty quantification. The posterior of MOGP enables seamless interpolation at any unseen index, making RR-FBTC well-suited for data with real-valued indices.

\textbf{Remark 3}: The algorithm requires an initial rank $R_{\text{init}}$, which influences the computational cost in the early stages. Domain knowledge can be used to select $R_{\text{init}}$ when available; otherwise, a practical heuristic is to set it to the maximum dimension of the tensor $\bc{Y}$. The algorithm is robust to $R_{\text{init}}$, and the rank decreases rapidly after initialization, significantly reducing computational costs, see App. C.

\section{Numerical Results}
\label{secV}
We evaluate the performance of the proposed RR-FBTC algorithm through extensive experiments on both synthetic and real-world datasets. 
All experiments are conducted in Matlab 2021b or later on a macOS system with an M1 CPU and 16 GB of RAM.

\subsection{Experimental settings}
\textbf{Evaluation metric}: For synthetic data, sound speed field (SSF) data, and US-Temperature data, we use root relative squared error (RRSE) and root mean squared error (RMSE): 
\begin{itemize}
	\item RRSE is defined as $\|\bc{X} - \hat{\bc{X}}\|_F/\|\bc{X}\|_F$, where $\bc{X}$ represents the ground-truth and $\hat{\bc{X}}$ is the reconstructed tensor.
	\item RMSE is defined as $\|\bc{X} - \hat{\bc{X}}\|_F/ \sqrt{T} $, where $T$ is the total number of entries in the tensor.
\end{itemize}
For image data, peak-signal-to-noisy ratio (PSNR) \cite{1284395} and structural similarity (SSIM) \cite{1284395} are used as performance metrics. 

\textbf{Baseline algorithms}: 
To validate the effectiveness of RR-FBTC, we compare it with several state-of-the-art low-rank tensor-based methods, FTNN \cite{jiang2020framelet}, a LRTC method with total variation regularization (LRTC-TV \cite{jiang2018anisotropic}), the Bayesian CP decomposition method FBCP \cite{zhao2015bayesian}, and the neural network-based continuous tensor decomposition method LRTFR \cite{luo2023low}. In all experiments, we use the default settings for hyperparameters and network architectures as suggested in the original papers.
The kernel used in RR-FBTC is the Matern kernel \cite{genton2001classes} with the degree of $\frac{5}{2}$, which is a widely adopted kernel with high expressive power. It is defined as
\begin{align}
    \varsigma\left(l ; h\right)=\bigg(1+\frac{\sqrt{5} l}{h}+
    \frac{5 l^{2}}{3 h^{2}}\bigg) \exp \bigg(-\frac{\sqrt{5} l}{h}\bigg),
\end{align}
where $l$ denotes the absolute distance between two indices, and $h$ is the kernel length scale, selected from the set $\{0.5, 1, 5, 10\}$ in our experiments. 
An ablation study on the involved kernel parameters shows the robustness of RR-FBTC to kernel parameter choices, see App. D.

\subsection{Synthetic data}
Two synthetic datasets are generated to evaluate RR-FBTC in terms of tensor rank estimation, reconstruction, and the ability to handle tensors with real-valued indices.
Accordingly, this section focuses on comparisons with SOTA methods that can either learn the tensor rank (FBCP) or reconstruct continuous tensors (LRTFR and FunBaT). Comparisons with other methods are presented in Sec.~\ref{secV.D} using real-world data.
All results in this subsection are averaged over ten Monte Carlo trials.

\begin{table}[t]
    \centering
    \caption{[Synthetic discrete data] RRSEs with standard deviations for synthetic data at different SNRs and sampling rates (SR). LRTFR-CP uses the ground-truth rank, while FBCP and RR-FBTC automatically infer it.}
    \begin{tabular}{ccccc}
        \toprule
        SR & SNR & FBCP & LRTFR-CP & RR-FBTC \\
        \midrule
        \multirow{4}{*}{20\%} 
        & 10 dB & 0.140$\pm$.007  & 0.141$\pm$.002 & \textbf{0.134}$\pm$.002 \\
        & 5 dB  & 0.251$\pm$.010 & 0.249$\pm$.003 & \textbf{0.239}$\pm$.004 \\
        & 0 dB  & 0.626$\pm$.028 & 0.617$\pm$.008 & \textbf{0.548}$\pm$.013 \\
        & -5 dB & 0.891$\pm$.032 & 0.884$\pm$.011 & \textbf{0.877}$\pm$.018 \\
        \midrule
        \multirow{4}{*}{30\%} 
        & 10 dB & 0.109$\pm$.005 & 0.108$\pm$.002 & \textbf{0.107}$\pm$.001 \\
        & 5 dB  & 0.189$\pm$.008 & 0.187$\pm$.003 & \textbf{0.182}$\pm$.004 \\
        & 0 dB  & 0.434$\pm$.022 & 0.371$\pm$.007 & \textbf{0.346}$\pm$.010 \\
        & -5 dB & 0.785$\pm$.029 & 0.809$\pm$.010 & \textbf{0.674}$\pm$.015 \\
        \bottomrule
    \end{tabular}
    \label{tab: sysdata}
\end{table}

\begin{figure}[t]
	\centering
	\includegraphics[width=0.80\linewidth]{fig/rankrrse.pdf}
	\caption{[Synthetic discrete data] Tensor rank estimates and RRSEs of the proposed RR-FBTC and FBCP under different SNRs. The blue dashed line is the ground-truth rank and the error bars show the standard deviation.}
        \label{fig: tre}
\end{figure}

We first consider a three dimensional discrete tensor $\bc{X}=\llbracket \mb{U}^1, \mb{U}^2, \mb{U}^3 \rrbracket\in \mathbb{R}^{30\times 30\times 30}$ with a rank of 10. Each element of the factor matrix $\mb{U}^k$ is independently drawn from a standard Gaussian distribution $\mathcal{N}(0, 1)$. 
The data are partially missing and corrupted by i.i.d. Gaussian noise with a variance of $\sigma_y^2$.
Specifically, the observed data is modeled as $\bc{Y} = \bc{O}\circledast( \bc{X} + \bc{W}) $ with $\bc{W}_{i,j,k}\sim \mathcal{N}(0, \sigma_y^2)$. 
We compare RR-FBTC with the Bayesian tensor completion method FBCP, initializing the rank in both FBCP and RR-FBTC to the maximum dimension size, i.e., $R_{\text{init}}=30$. 
Additionally, we consider the CP form (\ref{eq:7}) of the network-based method LRTFR, denoted as LRTFR-CP, setting its rank to the ground truth.

Table \ref{tab: sysdata} presents the RRSEs with standard deviations under varying sampling rates (SRs) and signal-to-noise ratios (SNRs), where the SNR is defined as 
$\mathrm{SNR} = 10 \log_{10}\left(\frac{\|\bc{X}_F\|^2}{T\sigma^2}\right)$ \cite{liu2012tensor}.
More results are in App. E.
It can be seen that the RRSEs of all methods decrease as SNR and SR increase. The RRSE for FBCP increases as the noise level rises. 
The estimated ranks and RRSEs of the Bayesian methods versus SNR under an SR of $30\%$ are shown in Fig.~\ref{fig: tre}.
Both methods provide accurate rank estimates at high SNR. At low SNR (e.g., $-5$ dB), FBCP's performance deteriorates, leading to lower rank estimates with high variability. In contrast, RR-FBTC exhibits greater robustness, maintaining rank estimates closer to the true (dashed line) even under challenging low-SNR conditions. Further discussions on Bayesian tensor completion in low-SNR scenarios is in \cite{cheng2022rethinking}.

\begin{figure}[t]
	\centering
	\includegraphics[width=0.80\linewidth]{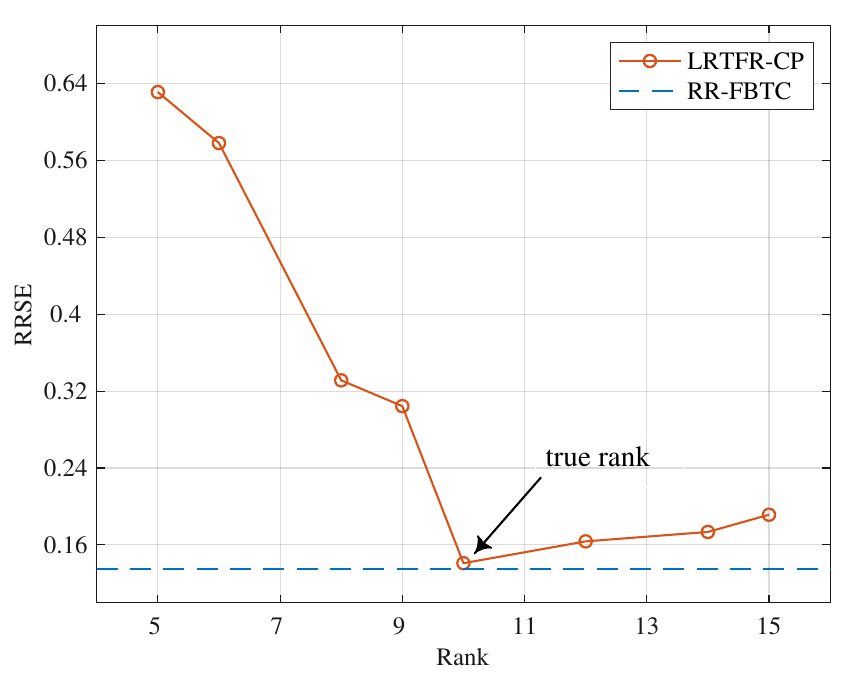}
	\caption{[Synthetic discrete data] RRSE of LRTFR with different rank settings. The blue dashed line is the RRSE of RR-FBTC.}
        \label{fig: lrtfr}
\end{figure}

RR-FBTC consistently outperforms LRTFR-CP across all scenarios, even when LRTFR-CP utilizes the ground truth rank. This is attributed to that the LRTFR-CP does not explicitly account for noise, which cause overfitting of the neural network. Additionally, the RRSEs of LRTFR-CP with varying tensor ranks show that the performance of LRTFR-CP is highly sensitive to the choice of rank, see Fig.~\ref{fig: lrtfr}.

Next, we consider a tensor sampled from a continuous function defined as follows:
\begin{align}
	f(x, y, z) = \mb{U}^1(x)\mb{U}^2(y)\mb{U}^3(z),
\end{align}
where $\mb{U}^1(x)=\sin^2 (2\pi x) \cos(2\pi x)$, $\mb{U}^2(y)=\sin(\frac{\pi}{4}y)(1-\sin^3(\frac{\pi}{2}y))$, and $\mb{U}^3(z)=\exp(-2z)\sin(\frac{3}{2}\pi z)$. 
We randomly sample real-valued coordinates $\{\mathcal{S}_1, \mathcal{S}_2, \mathcal{S}_3\}$ from the range $[0, 1]$ and construct a tensor $\bc{Y}\in \mathbb{R}^{50\times 50\times 50}$. 
Observations are obtained by sampling from the noisy data with SNR = 10 dB. 
Given that the tensor is the product of vectors, its rank is 1.
Standard low-rank tensor-based methods cannot be directly applied to this continuous data.
Therefore, we focus on comparing RR-FBTC with LRTFR-CP and FunBaT.

\begin{figure}[t]
	\centering
	\includegraphics[width=0.95\linewidth]{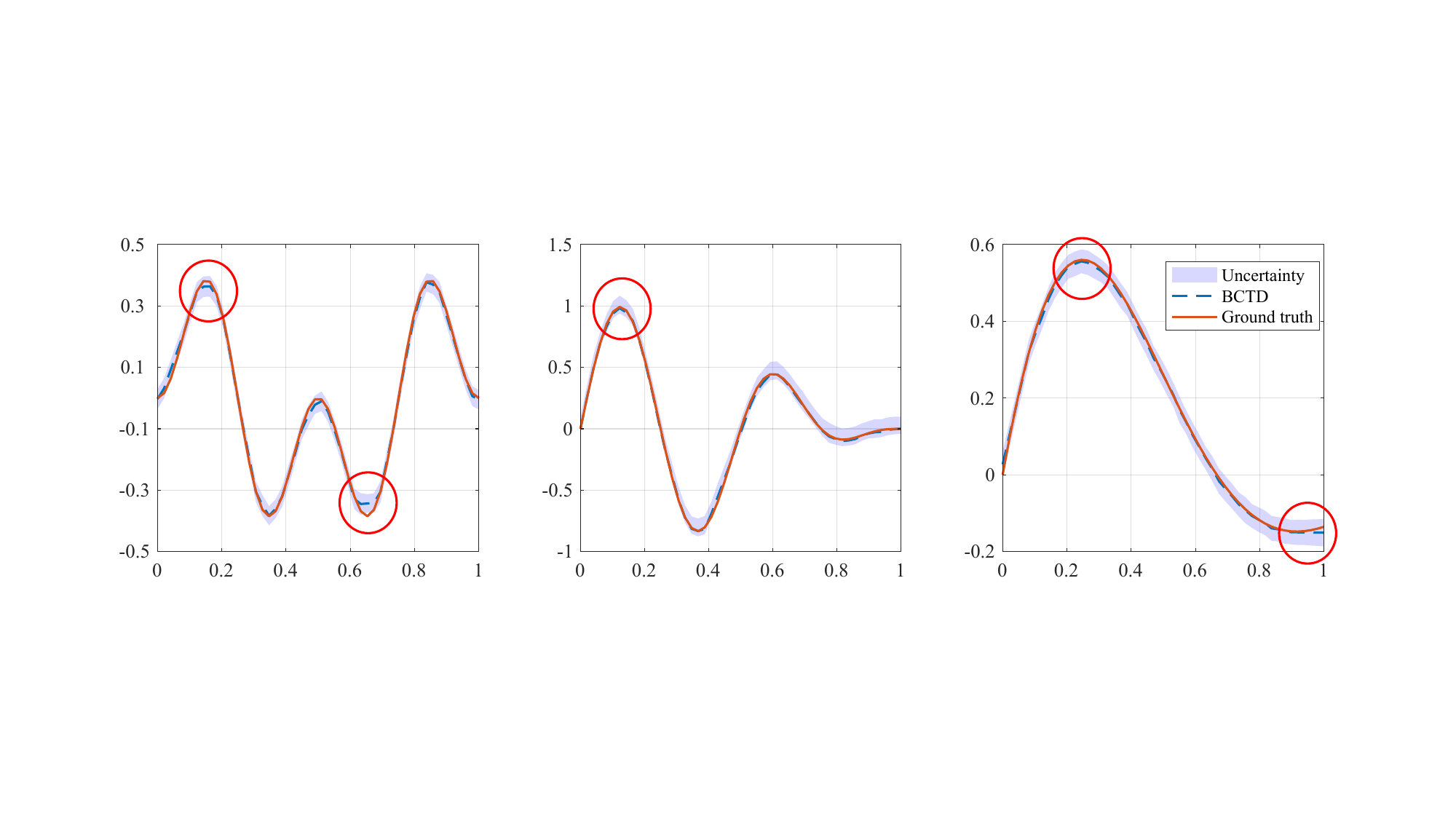}
	\caption{[Synthetic continuous data] The estimated latent factors and respective uncertainty of RR-FBTC for the synthetic continuous data. The SR = 20\% and SNR = 10 dB. }
        \label{fig: sfunc}
\end{figure}

As shown in Table \ref{tab: sytdata}, RR-FBTC can effectively handle tensors with real-valued indices, yielding lower RRSEs than  both LRTFR-CP and FunBaT.
Notably, LRTFR-CP and FunBaT perform well only when the true rank is correctly specified (rank 1 in this case). When the rank is overestimated (e.g., rank 3), the performance of LRTFR-CP and FunBaT deteriorates. In contrast, RR-FBTC automatically learns the rank and produces robust results, demonstrating its superiority in scenarios with real-valued indices and noisy observations. 
The learned latent functions alongside their ground truth indicate that RR-FBTC infers the correct rank and accurately identifies the latent functions, see Fig. \ref{fig: sfunc}. 
The proposed RR-FBTC identifies the true interpretable factors, which are compositions of trigonometric and exponential functions, even under noise and sparsity. This confirms that our model learns meaningful latent structures from the observed data.  
The estimated uncertainties in Fig. \ref{fig: sfunc} are characterized by the standard deviation of the learned factors. Regions with higher uncertainty (red circles) correspond to less accurate predictions, confirming the validity of the uncertainty estimates.

\begin{table}[t]
    \centering
    \caption{[Synthetic continuous data] RRSEs with standard deviations of LRTFR-CP and RR-FBTC with SRs at SNR=10 dB, ranks is in brackets.}
	\begin{tabular}{cccc}
	\toprule
	SR & LRTFR-CP & FunBaT & RR-FBTC \\ 
	\midrule
	\multirow{2}{*}{10\%} & 0.037$\pm$.004 (1) & 0.042$\pm$.005 (1) & \multirow{2}{*}{\textbf{0.035$\pm$.003} (1)} \\ 
	                      & 0.073$\pm$.006 (3) & 0.055$\pm$.006 (3) &            \\ 
	\midrule
	\multirow{2}{*}{20\%} & 0.027$\pm$.004 (1) & 0.035$\pm$.005 (1) & \multirow{2}{*}{\textbf{0.024$\pm$.003} (1)} \\ 
                      & 0.051$\pm$.005 (3) & 0.046$\pm$.005 (3) &            \\ 
	\bottomrule
	\end{tabular}
    \label{tab: sytdata}
\end{table}	

\subsection{Real-world continuous data}
To demonstrate the effectiveness of RR-FBTC for continuous-indexed tensors, we evaluate its performance on the {\it US-Temperature} data, obtained from the ClimateChange\footnote{https://berkeleyearth.org/data/}. 
This dataset records temperatures from cities worldwide along with geospatial features. We selected temperatures from 248 cities in the United States from 1750--2016, represented as a tensor with three continuous-indexed modes: latitude, longitude, and time. 
The strong spatiotemporal correlations in temperature data result in a low-rank tensor structure.
The tensor contains 20k observations, of which 18k are used for training and 2k for testing.
Notably, the training data account for only $4.7\%$ of the total tensor entries in $\bc{Y}$, highlighting the difficulty of the reconstruction task.

Table~\ref{tab: ustemp} presents the RMSEs of different continuous tensor methods. For LRTFR-CP, performance is evaluated under different rank settings since the ground-truth rank is unknown. The results show that RR-FBTC consistently outperforms LRTFR-CP and FunBaT across all rank configurations, demonstrating its ability to handle continuous-indexed data. Additionally, RR-FBTC provides accurate rank estimation automatically, eliminating manual tuning.

\begin{figure*}[t]
	\centering
	\includegraphics[width=0.85\linewidth]{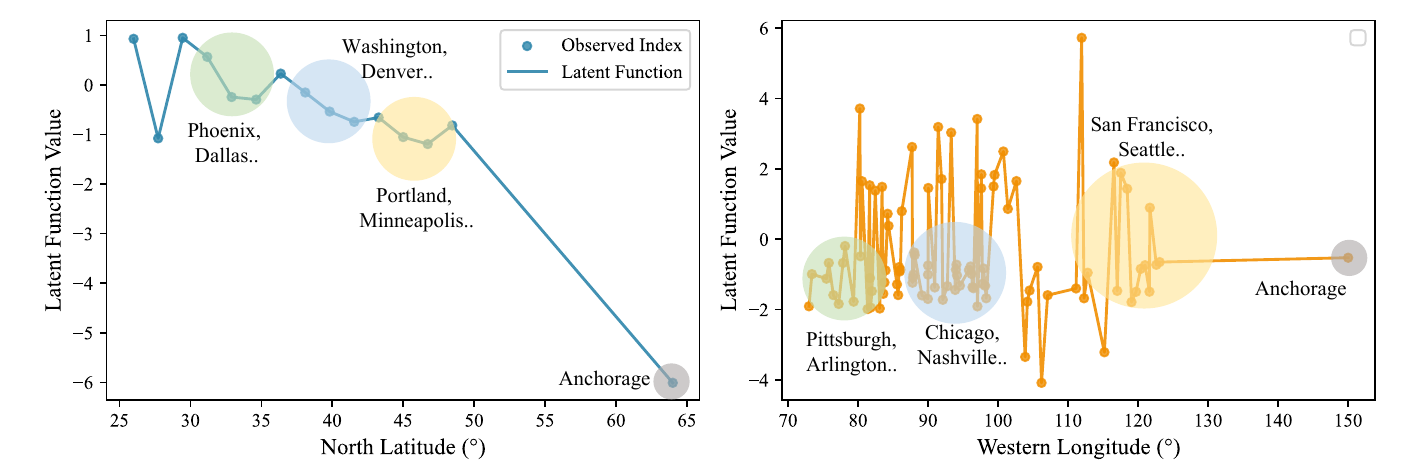}
	\caption{[{\it US-Temperature} data] Learned latent factors, left: corresponds to $\mathbf{U}^1$ (latitude) and right: corresponds to $\mathbf{U}^2$ (longitude).}
	\label{fig:usfact}
\end{figure*}

Moreover, we present an interpretability analysis of the \textit{US-Temperature} data. Specifically, we analyze the data using a rank-1 decomposition: $\bc{X}=\llbracket \mb{U}^1, \mb{U}^2, \mb{U}^3\rrbracket$, where each factor is a 1d vector. The resulting factors ($\mathbf{U}^1$ and $\mathbf{U}^2$) for the latitude and longitude modes are in Fig.~\ref{fig:usfact}, where the latent dimensions map directly to real geographic coordinates. Groups of cities are highlighted with shaded circles for clarity. 
 The latitude factor (left) exhibits a clear gradient from south to north: high factor values correspond to southern cities like Miami, Florida, while lower values align with northern locations such as Anchorage, Alaska. This reflects the well-known latitudinal decline in average temperature. Similarly, the longitude factor (right) reveals a smooth transition from east to west, capturing more subtle regional climate variations across the continent.
These decompositions not only validate the model’s ability to recover physically meaningful patterns but also offer insight into the dominant geographic trends governing temperature variation.

\begin{table}[t]
    \centering
    \caption{[{\it US-Temperature} data] RMSEs with standard deviations of the RR-FBTC and LRTFR-CP.}
	\begin{tabular}{lcc}
	\toprule
	Method & RMSE & Rank \\ 
	\midrule
    \multirow{3}{*}{LRTFR-CP} & 0.388$\pm$0.023  & 3 \\ 
                               & 0.354$\pm$0.015  & 5 \\ 
                               & 0.370$\pm$0.018  & 8 \\
	\midrule
    \multirow{3}{*}{FunBaT} & 0.805$\pm$0.014  & 3 \\ 
                               & 0.548$\pm$0.020  & 5 \\ 
                               & 0.551$\pm$0.021  & 7 \\
	\midrule
	RR-FBTC  & \textbf{0.342$\pm$0.012} & 5 (learned) \\ 
	\bottomrule
	\end{tabular}
	\label{tab: ustemp}
\end{table}

\subsection{Real-world on-grid data}
\label{secV.D}
To facilitate the comparison of RR-FBTC with non-functional tensor completion approaches, we evaluated its performance on two real-world on-grid datasets.

First, we utilize the three-dimensional (3D) sound speed field (SSF) \cite{li2023striking, cheng2022tensor}.
The SSF data, sized at $20\times20\times20$, is derived from the conductivity, temperature, and depth measurements using the hybrid coordinate ocean model\footnote{https://www.hycom.org}. 
The dataset corresponds to a region in the South China Sea, with specific longitude and latitude provided in \cite{cheng2022tensor}. It contains the sound speed values in a 3D geographical region, which covers a spatial area of 152 km $\times$ 152 km $\times$ 190 m (horizontal resolution 8 km, vertical resolution 10 m).
Observations are generated by adding i.i.d. Gaussian noise to the data and randomly sampling from it.
For LRTFR, the multi-linear rank is set to $[10, 10, 10]$.
For both FBCP and RR-FBTC, the upper rank bound is set to the maximum value of the dimensions.
A more accurate reconstruction of the SSF enables a better characterization of sound propagation.

\begin{figure*}[t]
	\centering
	\includegraphics[width=0.9\linewidth]{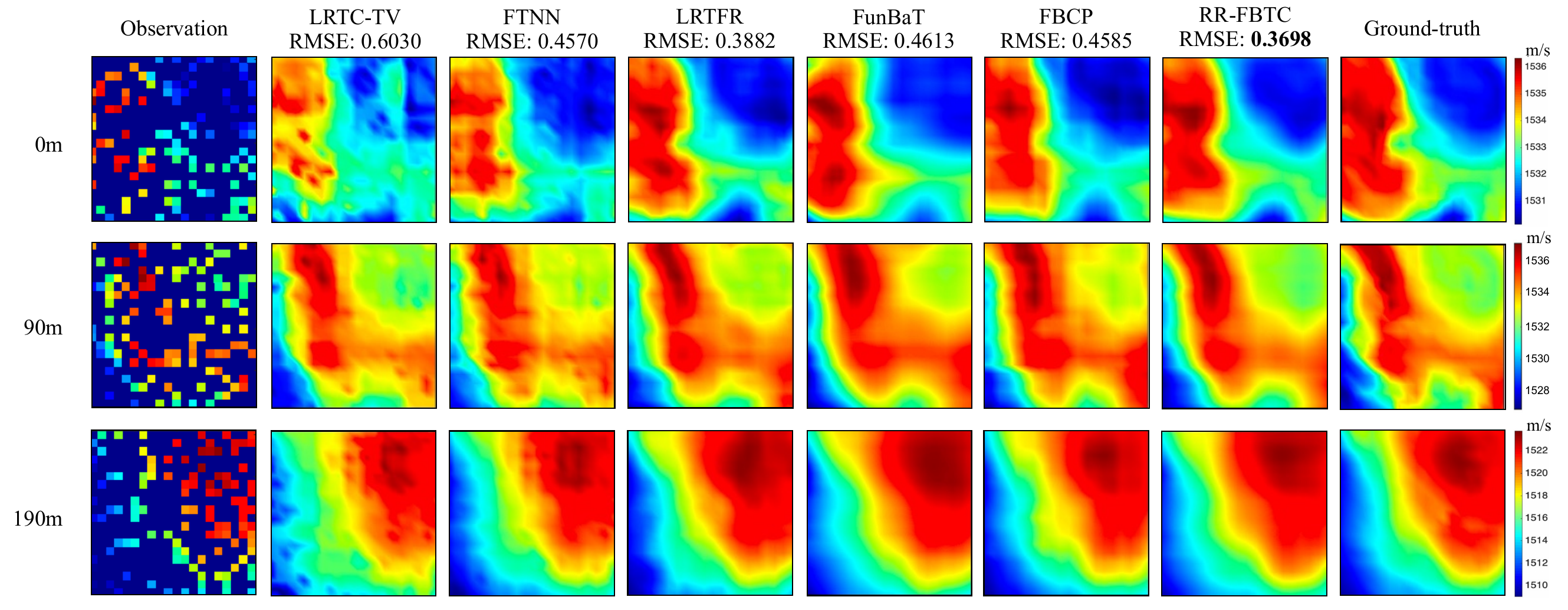}
	\caption{[SSF data] The reconstructed SSF of different methods in one Monte Carlo trial. The SR$=30\%$ and SNR = 20 dB.}
        \label{fig: ssfdata}
\end{figure*}

The quantitative and qualitative results of 3D SSF reconstruction are in Table \ref{tab: ssfdata} and Fig.~\ref{fig: ssfdata}.
The RR-FBTC achieves the best RMSEs and visual qualities in all scenarios.
The promising results of RR-FBTC is attributed to the concurrent modeling of the low-rankness and continuity of the data.
Moreover, the performance of RR-FBTC and FunBaT surpass that of the carefully designed total-variation-based method LRTC-TV, highlighting the advantages of continuous modeling in promoting smoothness. At high SNR and SR, LRTFR shows an RMSE close to that of RR-FBTC, which is lower than that of FBCP. However, as SNR decreases, the RMSE for LRTFR increases due to the lack of noise modeling.

\begin{table*}[ht]
    \centering
    \caption{[SSF data] RMSEs of different methods under varying SRs and SNRs, averaged over ten Monte Carlo trials.}
    \begin{tabular}{cccccccc}
        \toprule
        SR & SNR & LRTC-TV & FTNN & LRTFR & FunBaT & FBCP & RR-FBTC \\
        \midrule
        \multirow{2}{*}{10\%} & 20 dB & 1.305 & 0.866 & 0.893 & 0.704 & 0.697 & \textbf{0.475} \\
            & 10 dB & 1.449 & 1.112 & 1.362 & 0.995  &  0.961 & \textbf{0.597} \\
        \multirow{2}{*}{20\%} & 20 dB & 0.812 & 0.561 & 0.453 & 0.496  & 0.527 & \textbf{0.392} \\
            & 10 dB & 1.065 & 0.898 & 0.793 & 0.754 & 0.787 & \textbf{0.556} \\
        \multirow{2}{*}{30\%} & 20 dB & 0.603 & 0.456 & 0.374 & 0.466  & 0.458 & \textbf{0.358} \\
            & 10 dB & 0.923 & 0.840 & 0.665 & 0.701 & 0.604 & \textbf{0.519} \\
        \bottomrule
    \end{tabular}
    \label{tab: ssfdata}
\end{table*}

We evaluate the RR-FBTC on the image inpainting task with 8 RGB benchmark images, where each image is represented by a third-order tensor of size $256\times256\times3$.
The multilinear rank for LRTFR method is $[128, 128, 3]$.
For FBCP and RR-FBTC, the upper bound of rank is 200.
The SNR is $20$ dB.

\begin{figure*}[t]
	\centering
	\includegraphics[width=0.9\linewidth]{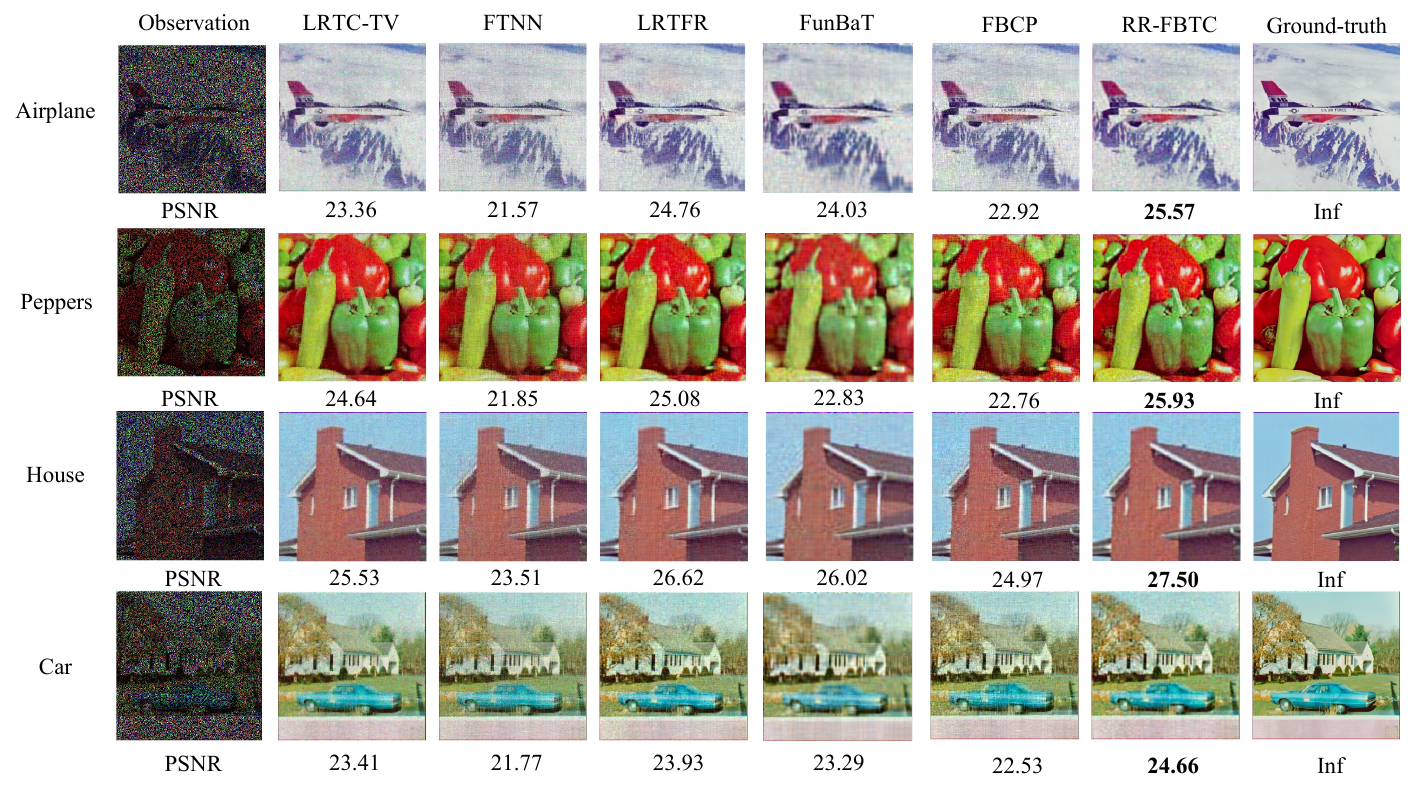}
	\caption{[Image data] Visual effects of the reconstructed images of different methods in one Monte Carlo trial. The SR = 30\% and SNR = 20 dB.}
        \label{fig: images}
\end{figure*}

The reconstruction results of different SOTA methods under varying SRs are in Table~\ref{tab: image}. 
It can be seen that the RR-FBTC is the most stable method among the algorithms in terms of different SRs and images.
The Bayesian methods, such as FBCP and RR-FBTC, outperform the low-rank-based method FTNN. This is attributed to the automatic model selection and robustness to noise of Bayesian methods.
LRTFR and FunBaT generally achieve the second- and third-best performance across most scenarios, highlighting the benefit of continuous modeling for the image inpainting task.
To examine the visual differences, four example images and the corresponding results are in Fig.~\ref{fig: images}.
The results show that RR-FBTC recovers the best images and loses fewer image details than LRTC-TV, FTNN, FunBaT and FBCP, revealing the superiority of RR-FBTC in handling on-grid data with integer indices.

\begin{table*}[ht]
    \centering
    \caption{[Image data] PSNRs and SSIMs of image completion at different SRs at SNR = 20 dB. Averaged over ten Monte Carlo trials.}
    \begin{tabular}{cccccccccccccccc}
        \toprule
        \multirow{2}{*}{SR} & \multirow{2}{*}{Image} & \multicolumn{2}{c}{LRTC-TV} & \multicolumn{2}{c}{FTNN} & \multicolumn{2}{c}{LRTFR} & \multicolumn{2}{c}{FunBaT} & \multicolumn{2}{c}{FBCP} & \multicolumn{2}{c}{RR-FBTC} \\
        & & RSNR & SSIM & RSNR & SSIM & RSNR & SSIM & RSNR & SSIM & RSNR & SSIM & RSNR & SSIM \\ 
        \midrule
        \multirow{8}{*}{20\%}
        & Peppers & 23.12 & 0.919 & 19.65 & 0.835 & 23.31 & 0.919 & 22.11 & 0.902 & 20.74 & 0.863 & \textbf{25.01} & \textbf{0.943} \\ 
        & Car & 22.45 & 0.748 & 20.45 & 0.632 & 22.67 & 0.761 & 22.78 & 0.780 & 20.99 & 0.684 & \textbf{23.53} & \textbf{0.820} \\ 
        & Barbara & 23.85 & 0.797 & 21.32 & 0.687 & 24.32 & 0.799 & 23.85 & 0.789 & 21.95 & 0.703 & \textbf{24.85} & \textbf{0.819} \\ 
        & House & 24.72 & 0.856 & 22.09 & 0.757 & 25.57 & 0.880 & 25.01 & 0.873 & 22.84 & 0.794 & \textbf{26.83} & \textbf{0.904} \\ 
        & Airplane  & 22.58 & 0.554 & 20.35 & 0.414 & 23.62 & 0.635 & 23.11 & 0.454 & 20.99 & 0.458 & \textbf{24.62} & \textbf{0.711} \\ 
        & Sailboat & 22.04 & 0.790 & 19.81 & 0.651 & 22.64 & 0.807 & 21.66 & 0.784 & 20.77 & 0.714 & \textbf{23.06} & \textbf{0.838} \\ 
        & Baboon & 21.14 & 0.695 & 19.54 & 0.629 & 20.68 & 0.642 & 20.81 & 0.640 & 19.64  & 0.621 & \textbf{21.51} & \textbf{0.688} \\
        & Facade & 24.65 & 0.758 & 24.96 & 0.778 & 25.62 & 0.806  & 26.11 & 0.823 & 25.34 & 0.795 & \textbf{26.23} & \textbf{0.825} \\ 
        \midrule
        \multirow{8}{*}{30\%}
        & Peppers & 24.64 & 0.938 & 21.85 & 0.890 & 25.16 & 0.945 & 22.90 & 0.913 & 22.76 & 0.907 & \textbf{25.93} & \textbf{0.953} \\ 
        & Car & 23.43 & 0.770 & 21.78 & 0.690 & 23.91 & 0.804 & 23.26 & 0.818 & 22.53 & 0.749 & \textbf{24.65} & \textbf{0.847} \\ 
        & Barbara & 25.16 & 0.827 & 23.12 & 0.7571 & 25.36 & 0.831 & 24.52 & 0.808 & 24.03 & 0.784 & \textbf{26.18} & \textbf{0.852} \\ 
        & House & 25.53 & 0.872 & 23.52 & 0.806 & 26.69 & 0.903 & 26.05 & 0.893 & 24.97 & 0.852 & \textbf{27.50} & \textbf{0.916} \\ 
        & Airplane & 23.36 & 0.557 & 21.58 & 0.466 & 24.71 & 0.697 & 24.07 & 0.715 & 22.92 & 0.550 & \textbf{25.55} & \textbf{0.743} \\ 
        & Sailboat & 23.25 & 0.825 & 21.36 & 0.730 & 23.93 & 0.848 & 22.63 & 0.815 & 22.50 & 0.789 & \textbf{24.06} & \textbf{0.866} \\ 
        & Baboon & \textbf{22.10} & \textbf{0.752} & 20.76 & 0.705 & 21.27 & 0.663 & 21.76 & 0.710 & 20.77 & 0.690 & 22.00 & 0.739 \\
        & Facade & 25.65 & 0.800 & 25.71 & 0.806 & 26.24 & 0.829 & 27.07 & 0.848 & 26.69 & 0.840 & \textbf{27.52} & \textbf{0.860} \\  
        \bottomrule
    \end{tabular}
    \label{tab: image}
\end{table*}

\section{Conclusion}
This paper presented a rank-revealing functional Bayesian tensor completion method (RR-FBTC) that assigns MOGP priors to latent functions. This approach enables effective handling of tensors with real-valued indices and automatic determination of tensor rank during the inference procedure. Our theoretical analysis demonstrated the model's universal approximation property (UAP) for any multi-dimensional continuous signals, highlighting its high expressive power while maintaining mathematical conciseness. Furthermore, we showed that a number of existing matrix/tensor decomposition methods are special cases of RR-FBTC, offering insight into its properties.
Based on the variational inference framework, we derived an efficient algorithm with closed-form updates. 
Extensive experiments on synthetic and real-world datasets demonstrated the method's effectiveness and superiority, especially for low SNR and sparse observations.

\appendices
\section{Proofs}
\numberwithin{equation}{section}
\setcounter{equation}{0} 
\label{appenxid-a}
\subsection{Proof of Lemma 1}
Let $\varsigma$ be a universal kernel on the compact metric space $\mathcal{Z}$. By the definition of a universal kernel, the RKHS $\mathcal{H}$, corresponding to $\varsigma$, is dense in the space of continuous functions $C(\mathcal{Z})$ with respect to the norm $\|\cdot\|_{\infty}$.
From the construction of the RKHS $\mathcal{H}$, any function $f\in \mathcal{H}$ can be expressed as a limit of functions in $\mathcal{H}_0$, where $\mathcal{H}_0$ consists of finite linear combinations of kernel functions:
\begin{align}
	f(\cdot) = \sum_{n=1}^N c_i \varsigma(\cdot, \mb{z}_i), c_i \in \mathbb{R}, \mb{z}_i \in \mathcal{Z}, \forall i.
\end{align}
Thus, for any $f\in \mathcal{H}$, there exists a sequence of functions $f_k \in \mathcal{H}_0$, such that
\begin{align}
	\|f_k-f\|_\infty \rightarrow 0, \text{as}~k\rightarrow \infty.
\end{align}
Since $\mathcal{H}$ is dense in $C(\mathcal{Z})$, for any $g \in C(\mathcal{Z})$ and $\epsilon>0$, there exits an $f \in \mathcal{H}$ such that 
\begin{align}
	\|f - g\|_\infty < {\epsilon}{/2}.
\end{align} 
Because $f\in \mathcal{H}$, it can be approximated by $f_k \in \mathcal{H}_0$ (a finite linear combination of kernel functions) such that
\begin{align}
	\|f_k-f\|_\infty < {\epsilon}/{2}.
\end{align}
Let $f_k(\cdot)= \sum_{n=1}^N c_i \varsigma(\cdot, \mb{z}_i)$, where $\{c_i\}_{i=1}^N$ are coefficients and $\{\mb{z}_i\}_{i=1}^N$ are data points in $\mathcal{Z}$. Combining the two approximations above, we have
\begin{align}
	\|f_k - g\|_\infty \leq \|f_k-f\|_\infty + \|f - g\|_\infty < \epsilon.
\end{align}
Therefore, for any $g \in C(\mathcal{Z})$ and $\epsilon>0$, there exits coefficients $\{c_i\}_{i=1}^N$ and data points $\{\mb{z}_i\}_{i=1}^N$ such that
\begin{align}
	\|\sum_{n=1}^N c_i \varsigma(\cdot, \mb{z}_i) - g\|_\infty < \epsilon.
\end{align} 

\subsection{Proof of Theorem 1}
The posterior mean function $f(\mb{z})$ follows the CP form, allowing it to be expressed as
\begin{align}
	\label{eq: t1f}
	f(\mb{z}) = \sum_{r=1}^R\prod_{w=1}^D \bar{{u}}_r^w(z_w),
\end{align}
where $R$ is the tensor rank, $\mb{z}=[z_1,\cdots,z_D]^T$ and $\bar{{u}}_r^w$ represents the posterior mean of the latent factors.
Since each latent function ${u}_r^w(\cdot)$ follows a GP prior, its posterior mean is 
\begin{align}
	\label{eq: t1u}
	\bar{{u}}_r^w(z_w)=\sum_{k_d=1}^{N_d} a_w^r(k_d) \varsigma(z_w, y_{k_d}),
\end{align}
where $a_w^r \in \mathbb{R}$ are coefficients and $\{y_{k_d}\}_{k_d=1}^{N_d}$ are data values in the $w$-th dimension, for $w=1,\cdots,D$. 
Substituting (\ref{eq: t1u}) into (\ref{eq: t1f}), we obtain
\begin{align}
	f(\mb{z}) &= \sum_{r=1}^R\prod_{w=1}^D \sum_{k_d=1}^{N_d} a_w^r(k_d) \varsigma(z_w, y_{k_d})\nonumber \\
	&= \sum_{k_1=1}^{N_1} \cdots \sum_{k_D=1}^{N_D} \sum_{r=1}^R \prod_{w=1}^D a_w^r(k_d) \varsigma(z_w, y_{k_d}).
\end{align}
The second equation follows by interchanging the order of summation and product.
The coefficients $\{a_w^r\}$ can be seen as a rank-$R$ CP decomposition of the coefficient tensor $\mathcal{A}$, i.e.,
\begin{align}
	\mathcal{A}[k_1, \cdots, k_D]=\mathcal{A}_{\mb{k}} = \sum_{r=1}^R \prod_{w=1}^D a_w^r(k_d),
\end{align}
where $\mb{k}=[k_1,\cdots,k_D]^T$ is a multi-index.
Exploiting the factorization property of the tensor product kernel $\varsigma$, the function is reformulated as
\begin{align}
	\label{eq: t1fk}
	&f(\mb{z}) = \sum_{k_1=1}^{N_1} \cdots \sum_{k_D=1}^{N_D} \sum_{r=1}^R \prod_{w=1}^D a_w^r(k_d) \varsigma(z_w, y_{k_d}) \nonumber \\
	&\!\!\!\!\!= \sum_{k_1=1}^{N_1} \cdots \sum_{k_D=1}^{N_D} \mathcal{A}_{\mb{k}} \prod_{w=1}^D \varsigma(z_w, y_{k_d}) = \sum_{\mb{k}}\mathcal{A}_{\mb{k}} \varsigma(\mb{z}, \mb{y}_{\mb{k}}), 
\end{align}
where $\mb{y}_{\mb{k}}=[y_{k_1},\cdots,y_{k_D}]^T$. 
(\ref{eq: t1fk}) indicates that the posterior mean function is a linear combination of kernel functions. 
Since $\varsigma$ is a universal kernel, $f(\mb{z})$ can approximate any continuous function on the compact space $\mathcal{Z}$ to any desired accuracy based on \textbf{Lemma 1}.

\vspace{-0.3cm}
\section{Derivation of Algorithm 1}
\label{appenxid-b}
The logarithm of the joint probability density function $p(\mb{\Theta},\bc{Y})$ is
\begin{align}
	\label{eq: logjpdf}
	\text{ln}&p(\mb{\Theta},\bc{Y}) = \frac{N}{2}\ln \tau - \frac{\tau}{2}\bigg\|\bc{O}\circledast(\bc{Y}-\llbracket \mb{U}^1,\cdots,\mb{U}^K \rrbracket)\bigg\|_F^2 \nonumber \\
	&+\sum_{k=1}^K\sum_{r=1}^R\left[ \frac{N_k}{2}\ln \gamma_r-\frac{1}{2}\gamma_r (\mb{u}_r^k)^T\mb{\Sigma}_k^{-1}\mb{u}_r^k \right]\nonumber \\
	&+ \sum_{r=1}^R\left[(a_r-1)\text{ln}\gamma_r - b_r\gamma_r\right]+(a_0-1)\ln\tau -b_0\tau + \text{const}.
\end{align}
We omit the constant in the following.
Exploiting the mean-field assumption, the variation distribution is factorized as
\begin{align}
\!\!\!\!\!q(\mb{\Theta})=q(\boldsymbol{\gamma})q(\tau)\prod_{k=1}^K q(\mb{U}^k) =q(\boldsymbol{\gamma})q(\tau)\prod_{k=1}^K \prod_{r=1}^Rq(\mb{u}_r^k).
\end{align} 
The optimal pdf for each variable group is obtained by substituting the logarithm of the joint probability distribution (\ref{eq: logjpdf}) into (28) of the main text.

Particularly, the logarithm of $q(\mb{u}_r^k)$ is given in (B.3),
\begin{figure*}[!htb]
\normalsize
\setcounter{equation}{2}
\begin{align}
	&\ln q(\mb{u}_r^k)=\mathbb{E}_{q(\mb{\Theta}\setminus \mb{u}_r^k)}\bigg\{ - \frac{\tau}{2}\bigg\|\bc{O}\circledast(\bc{Y}-\llbracket \mb{U}^1,\cdots,\mb{U}^K \rrbracket)\bigg\|_F^2 - \frac{1}{2}\gamma_r (\mb{u}_r^k)^T\mb{\Sigma}_k^{-1}\mb{u}_r^k \bigg\}\nonumber \\
	&=\mathbb{E}_{q(\mb{\Theta}\setminus \mb{u}_r^k)} \bigg\{ - \frac{\tau}{2}\bigg[(\mb{u}_r^k)^T \text{diag}[\bc{O}_{(k)} \bigodot_{\substack{l=K\\ l\neq k}}^1 (\mb{u}_r^l \circledast \mb{u}_r^l) ]\mb{u}_r^k - 2(\mb{u}_r^k)^T\bigg[\bc{O}\circledast(\bc{Y}-\sum_{\substack{s=1\\ s\neq r}}^R\mb{u}_s^1\circ\cdots\circ \mb{u}_s^K )\bigg]_{(k)}\bigodot_{\substack{l=K\\ l\neq k}}^1 \mb{u}_r^l \bigg] -\frac{1}{2}\gamma_r (\mb{u}_r^k)^T\mb{\Sigma}_k^{-1}\mb{u}_r^k \bigg\} \nonumber \\
	&=-\frac{1}{2} (\mb{u}_r^k)^T\bigg[ \lceil \tau \rceil \text{diag}\bigg[\bc{O}_{(k)} \bigodot_{\substack{l=K\\ l\neq k}}^1 \lceil\mb{u}_r^l \circledast \mb{u}_r^l\rceil \bigg] + \lceil\gamma_r\rceil\mb{\Sigma}_k^{-1} \bigg]\mb{u}_r^k + (\mb{u}_r^k)^T\bigg[ \lceil \tau \rceil \bigg[\bc{O}\circledast(\bc{Y}-\sum_{\substack{s=1\\ s\neq r}}^R\lceil\mb{u}_s^1\circ\cdots\circ \mb{u}_s^K \rceil)\bigg]_{(k)}\bigodot_{\substack{l=K\\ l\neq k}}^1 \lceil\mb{u}_r^l\rceil \bigg] \nonumber \\
	& =-\frac{1}{2} (\mb{u}_r^k)^T(\boldsymbol{\Psi}_r^k)^{-1}\mb{u}_r^k + (\mb{u}_r^k)^T(\boldsymbol{\Psi}_r^k)^{-1}\mb{m}_r^k, 
\end{align}
\hrulefill
\vspace*{4pt}
\end{figure*} 
where 
\begin{align}
	\mb{m}_r^k &= \lceil\tau\rceil \boldsymbol{\Psi}_r^k [\bc{O}\circledast(\bc{Y}-\sum_{\substack{s=1\\ s\neq r}}^R\lceil\mb{u}_s^1\circ\cdots\circ \mb{u}_s^K \rceil)]_{(k)}\bigodot_{\substack{l=K\\ l\neq k}}^1 \lceil\mb{u}_r^l\rceil,\\
	\boldsymbol{\Psi}_r^k &= \{ \lceil \tau \rceil \text{diag}[\bc{O}_{(k)} \bigodot_{\substack{l=K\\ l\neq k}}^1 \lceil\mb{u}_r^l \circledast \mb{u}_r^l\rceil ] + \lceil\gamma_r\rceil\mb{\Sigma}_k^{-1} \}^{-1}.
\end{align}
Thus, $q(\mb{u}_r^k)$ follows a Gaussian distribution
\begin{align}
	q(\mb{u}_r^k)=\mathcal{N}(\mb{u}_r^k|\mb{m}_r^k,\boldsymbol{\Psi}_r^k).
\end{align}
Similarly, the logarithm of $q(\boldsymbol{\gamma})$ can be formulated as in (B.7),
\begin{figure*}[!htb]
\vspace{-0.5cm}
\normalsize
\setcounter{equation}{6}
\begin{align}
	\ln &q(\boldsymbol{\gamma})=\mathbb{E}_{q(\mb{\Theta}\setminus \boldsymbol{\gamma})}\bigg\{\sum_{k=1}^K\sum_{r=1}^R\bigg[ \frac{N_k}{2}\ln \gamma_r-\frac{1}{2}\gamma_r (\mb{u}_r^k)^T\mb{\Sigma}_k^{-1}\mb{u}_r^k \bigg] + \sum_{r=1}^R\left[(a_r-1)\text{ln}\gamma_r - b_r\gamma_r\right] \bigg\} \nonumber \\
	&= \sum_{k=1}^K\sum_{r=1}^R\left[ \frac{N_k}{2}\ln \gamma_r-\frac{1}{2}\gamma_r \lceil (\mb{u}_r^k)^T\mb{\Sigma}_k^{-1}\mb{u}_r^k \rceil \right] + \sum_{r=1}^R\left[(a_r-1)\text{ln}\gamma_r - b_r\gamma_r\right] \} \nonumber \\
	& = \sum_{r=1}^R \bigg [ \bigg(a_r + \sum_{k=1}^K\frac{N_k}{2} -1 \bigg)\ln \gamma_r - \left( b_r +  \frac{1}{2}\gamma_r \lceil (\mb{u}_r^k)^T\mb{\Sigma}_k^{-1}\mb{u}_r^k \rceil \right)\gamma_r\bigg]	=\sum_{r=1}^R \big [\left(\hat{a}_r-1\right)\ln \gamma_r -\hat{b}_r\gamma_r\big ],
\end{align}
\hrulefill
\vspace*{4pt}
\end{figure*} 
where
\begin{align}
	\!\!\!\!\!\hat{a}_r = a_r+\frac{1}{2}\sum_{k=1}^K N_k,~~
	\hat{b}_r = b_r + \frac{1}{2}\sum_{k=1}^K\lceil (\mb{u}_r^k)^T\mb{\Sigma}_k^{-1}\mb{u}_r^k \rceil .
\end{align}
Thus, the optimal pdf for $\boldsymbol{\gamma}$ is a Gamma distribution, given by
\begin{align}
	q(\boldsymbol{\gamma})=\prod_{r=1}^R \text{Gam}(\gamma_r|\hat{a}_r,\hat{b}_r).
\end{align}

For parameter $\tau$, the logarithm of $q(\boldsymbol{\gamma})$ is
\begin{align}
	\ln &q(\tau)= \mathbb{E}_{q(\mb{\Theta}\setminus \boldsymbol{\gamma})} \bigg\{ \frac{N}{2}\ln\tau - \frac{\tau}{2}\|\bc{O}\circledast(\bc{Y}-\llbracket \mb{U}^1,\cdots,\mb{U}^K \rrbracket)\|_F^2 \nonumber \\ 
	&+(a_0-1)\ln\tau - b_0\tau \bigg\} \nonumber \\
	&= \big( \frac{N}{2} + a_0-1 \big) \ln \tau \nonumber \\
	&~~~ - \bigg[ \frac{1}{2} \lceil \left\| \bc{O}\circledast(\bc{Y}-\llbracket \mb{U}^1,\cdots,\mb{U}^K \rrbracket) \right\|_F^2 \rceil + b_0 \bigg] \tau\nonumber \\
	&=( \hat{a}_0-1 ) \ln \tau - \hat{b}_0 \tau,
\end{align}
where
\begin{align}
	\!\!\!\hat{a}_0\! = \!a_0+\frac{N}{2},\quad
	\hat{b}_0 \!= \!b_0+\frac{1}{2}\! \lceil \left\| \bc{O}\circledast(\bc{Y}-\llbracket \mb{U}^1,\cdots,\mb{U}^K \rrbracket) \!\right\|_F^2 \rceil.
\end{align}
Thus, the optimal pdf for $\tau$ is also a Gamma distribution
\begin{align}
	q(\tau) = \text{Gam}(\tau|\hat{a}_0,\hat{b}_0).
\end{align}

In the update of the optimal pdfs, there are several expectations to be computed. They can be obtained either from the statistical literature or similar results in related works \cite{tong2023bayesian, zhao2015bayesian}.

%
%
%
%
%


\bibliography{ref} 

@article{kolda2009tensor,
  title={Tensor decompositions and applications},
  author={Kolda, Tamara G and Bader, Brett W},
  journal={SIAM Review},
  volume={51},
  number={3},
  pages={455--500},
  year={2009},
  publisher={SIAM}
}

@article{bro1997parafac,
  title={PARAFAC. Tutorial and applications},
  author={Bro, Rasmus},
  journal={Chemom. Intell. Lab. Syst.},
  volume={38},
  number={2},
  pages={149--171},
  year={1997},
  publisher={Elsevier}
}

@article{oseledets2011tensor,
  title={Tensor-train decomposition},
  author={Oseledets, Ivan V},
  journal={SIAM J. Sci. Comput.},
  volume={33},
  number={5},
  pages={2295--2317},
  year={2011},
  publisher={SIAM}
}

@article{tucker1966some,
  title={Some mathematical notes on three-mode factor analysis},
  author={Tucker, Ledyard R},
  journal={Psychometrika},
  volume={31},
  number={3},
  pages={279--311},
  year={1966},
  publisher={Springer}
}

@article{zhao2016tensor,
  title={Tensor ring decomposition},
  author={Zhao, Qibin and Zhou, Guoxu and Xie, Shengli and Zhang, Liqing and Cichocki, Andrzej},
  journal={arXiv preprint arXiv:1606.05535},
  year={2016}
}

@article{chen2022reweighted,
  title={Reweighted low-rank factorization with deep prior for image restoration},
  author={Chen, Lin and Jiang, Xue and Liu, Xingzhao and Haardt, Martin},
  journal={IEEE Trans. Signal Process.},
  volume={70},
  pages={3514--3529},
  year={2022},
  publisher={IEEE}
}

@article{liu2012tensor,
  title={Tensor completion for estimating missing values in visual data},
  author={Liu, Ji and Musialski, Przemyslaw and Wonka, Peter and Ye, Jieping},
  journal={IEEE Trans. Pattern Anal. Mach. Intell},
  volume={35},
  number={1},
  pages={208--220},
  year={2012},
  publisher={IEEE}
}

@article{de2021channel,
  title={Channel estimation for intelligent reflecting surface assisted MIMO systems: A tensor modeling approach},
  author={de Ara{\'u}jo, Gilderlan T and de Almeida, Andr{\'e} LF and Boyer, R{\'e}my},
  journal={IEEE J. Sel. Topics Signal Process.},
  volume={15},
  number={3},
  pages={789--802},
  year={2021},
  publisher={IEEE}
}

@article{zhang2024integrated,
  title={Integrated sensing and communication with massive MIMO: A unified tensor approach for channel and target parameter estimation},
  author={Zhang, Ruoyu and Cheng, Lei and Wang, Shuai and Lou, Yi and Gao, Yulong and Wu, Wen and Ng, Derrick Wing Kwan},
  journal={IEEE Trans. Wirel. Commun.},
  volume={23},
  number={8},
  pages={8571--8587},
  year={2024},
  publisher={IEEE}
}

@article{li2023striking,
  title={Striking the right balance: Three-dimensional ocean sound speed field reconstruction using tensor neural networks},
  author={Li, Siyuan and Cheng, Lei and Zhang, Ting and Zhao, Hangfang and Li, Jianlong},
  journal={J. Acoust. Soc. Amer.},
  volume={154},
  number={2},
  pages={1106--1123},
  year={2023},
  publisher={AIP Publishing}
}

@article{shrestha2022deep,
  title={Deep spectrum cartography: Completing radio map tensors using learned neural models},
  author={Shrestha, Sagar and Fu, Xiao and Hong, Mingyi},
  journal={IEEE Trans. Signal Process.},
  volume={70},
  pages={1170--1184},
  year={2022},
  publisher={IEEE}
}

@article{zhao2015bayesian,
  title={Bayesian CP factorization of incomplete tensors with automatic rank determination},
  author={Zhao, Qibin and Zhang, Liqing and Cichocki, Andrzej},
  journal={IEEE Trans. Pattern Anal. Mach. Intell},
  volume={37},
  number={9},
  pages={1751--1763},
  year={2015},
  publisher={IEEE}
}

@article{cheng2022towards,
  title={Towards flexible sparsity-aware modeling: Automatic tensor rank learning using the generalized hyperbolic prior},
  author={Cheng, Lei and Chen, Zhongtao and Shi, Qingjiang and Wu, Yik-Chung and Theodoridis, Sergios},
  journal={IEEE Trans. Signal Process.},
  volume={70},
  pages={1834--1849},
  year={2022},
  publisher={IEEE}
}

@article{sidiropoulos2017tensor,
  title={Tensor decomposition for signal processing and machine learning},
  author={Sidiropoulos, Nicholas D and De Lathauwer, Lieven and Fu, Xiao and Huang, Kejun and Papalexakis, Evangelos E and Faloutsos, Christos},
  journal={IEEE Trans. Signal Process.},
  volume={65},
  number={13},
  pages={3551--3582},
  year={2017},
  publisher={IEEE}
}

@article{ten2002uniqueness,
  title={On uniqueness in CANDECOMP/PARAFAC},
  author={Ten Berge, Jos MF and Sidiropoulos, Nikolaos D},
  journal={Psychometrika},
  volume={67},
  pages={399--409},
  year={2002},
  publisher={Springer}
}

@inproceedings{fang2024functional,
  title={Functional {B}ayesian Tucker Decomposition for Continuous-indexed Tensor},
  author={Fang, Shikai and Yu, Xin and Wang, Zheng and Li, Shibo and Kirby, Robert M and Zhe, Shandian},
  year={2024},
  organization={Int. Conf. Learn. Represent. (ICLR)}
}

@article{luo2023low,
  title={Low-rank tensor function representation for multi-dimensional data recovery},
  author={Luo, Yisi and Zhao, Xile and Li, Zhemin and Ng, Michael K and Meng, Deyu},
  journal={IEEE Trans. Pattern Anal. Mach. Intell},
  year={2023},
  publisher={IEEE}
}

@article{chertkov2023black,
  title={Black box approximation in the tensor train format initialized by ANOVA decomposition},
  author={Chertkov, Andrei and Ryzhakov, Gleb and Oseledets, Ivan},
  journal={SIAM J. Sci. Comput.},
  volume={45},
  number={4},
  pages={A2101--A2118},
  year={2023},
  publisher={SIAM}
}

@article{hashemi2017chebfun,
  title={Chebfun in three dimensions},
  author={Hashemi, Behnam and Trefethen, Lloyd N},
  journal={SIAM J. Sci. Comput.},
  volume={39},
  number={5},
  pages={C341--C363},
  year={2017},
  publisher={SIAM}
}

@article{kargas2021supervised,
  title={Supervised learning and canonical decomposition of multivariate functions},
  author={Kargas, Nikos and Sidiropoulos, Nicholas D},
  journal={IEEE Trans. Signal Process.},
  volume={69},
  pages={1097--1107},
  year={2021},
  publisher={IEEE}
}

@article{cho2024separable,
  title={Separable physics-informed neural networks},
  author={Cho, Junwoo and Nam, Seungtae and Yang, Hyunmo and Yun, Seok-Bae and Hong, Youngjoon and Park, Eunbyung},
  journal={Adv. Neural Inf. Process. Syst.},
  volume={36},
  year={2024}
}

@inproceedings{schmidt2009function,
  title={Function factorization using warped {G}aussian processes},
  author={Schmidt, Mikkel N},
  booktitle={Int. Conf. Mach. Learn.},
  pages={921--928},
  year={2009}
}

@article{hornik1989multilayer,
  title={Multilayer feedforward networks are universal approximators},
  author={Hornik, Kurt and Stinchcombe, Maxwell and White, Halbert},
  journal={Neural Networks},
  volume={2},
  number={5},
  pages={359--366},
  year={1989},
  publisher={Elsevier}
}

@article{micchelli2006universal,
  title={Universal Kernels.},
  author={Micchelli, Charles A and Xu, Yuesheng and Zhang, Haizhang},
  journal={J. Mach. Learn. Res.},
  volume={7},
  number={12},
  year={2006}
}

@article{chen2023multivariate,
  title={Multivariate {G}aussian and Student-t process regression for multi-output prediction},
  author={Chen, Zexun and Wang, Bo and Gorban, Alexander N},
  journal={Neural Computing and Applications},
  volume={32},
  pages={3005--3028},
  year={2020},
  publisher={Springer}
}

@book{bishop2006pattern,
  title={Pattern recognition and machine learning},
  author={Bishop, Christopher M},
  year={2006},
  publisher={Springer}
}

@article{blei2017variational,
  title={Variational inference: A review for statisticians},
  author={Blei, David M and Kucukelbir, Alp and McAuliffe, Jon D},
  journal={J. Am. Stat. Assoc.},
  volume={112},
  number={518},
  pages={859--877},
  year={2017},
  publisher={Taylor \& Francis}
}

@article{ermis2014bayesian,
  title={A {B}ayesian tensor factorization model via variational inference for link prediction},
  author={Ermis, Beyza and Cemgil, A Taylan},
  journal={arXiv preprint arXiv:1409.8276},
  year={2014}
}

@article{song2023tensor,
  title={Tensor-based sparse {B}ayesian learning with intra-dimension correlation},
  author={Song, Yuhui and Gong, Zijun and Chen, Yuanzhu and Li, Cheng},
  journal={IEEE Trans. Signal Process.},
  volume={71},
  pages={31--46},
  year={2023},
  publisher={IEEE}
}

@article{chen2023bayesian,
  title={Bayesian low-rank matrix completion with dual-graph embedding: Prior analysis and tuning-free inference},
  author={Chen, Yangge and Cheng, Lei and Wu, Yik-Chung},
  journal={Signal Processing},
  volume={204},
  pages={108826},
  year={2023},
  publisher={Elsevier}
}

@article{aronszajn1950theory,
  title={Theory of reproducing kernels},
  author={Aronszajn, Nachman},
  journal={Trans. Am. Math. Soc.},
  volume={68},
  number={3},
  pages={337--404},
  year={1950}
}

@book{steinwart2008support,
  title={Support Vector Machines},
  author={Steinwart, Ingo},
  year={2008},
  publisher={Springer}
}

@article{kanagawa2018gaussian,
  title={Gaussian processes and kernel methods: A review on connections and equivalences},
  author={Kanagawa, Motonobu and Hennig, Philipp and Sejdinovic, Dino and Sriperumbudur, Bharath K},
  journal={arXiv preprint arXiv:1807.02582},
  year={2018}
}

@article{szabo2018characteristic,
  title={Characteristic and universal tensor product kernels},
  author={Szab{\'o}, Zolt{\'a}n and Sriperumbudur, Bharath K},
  journal={J. Mach. Learn. Res.},
  volume={18},
  number={233},
  pages={1--29},
  year={2018}
}

@article{jiang2020framelet,
  title={Framelet representation of tensor nuclear norm for third-order tensor completion},
  author={Jiang, Tai-Xiang and Ng, Michael K and Zhao, Xi-Le and Huang, Ting-Zhu},
  journal={IEEE Trans. Image Process.},
  volume={29},
  pages={7233--7244},
  year={2020},
  publisher={IEEE}
}

@inproceedings{jiang2018anisotropic,
  title={Anisotropic total variation regularized low-rank tensor completion based on tensor nuclear norm for color image inpainting},
  author={Jiang, Fei and Liu, Xiao-Yang and Lu, Hongtao and Shen, Ruimin},
  booktitle={IEEE Int. Conf. Acoust., Speech, Signal Process.},
  pages={1363--1367},
  year={2018}
}

@Inbook{Neal1996,
  author    = {Neal, Radford M.},
  title     = {Priors for Infinite Networks},
  booktitle = {Bayesian Learning for Neural Networks},
  year      = {1996},
  publisher = {Springer, New York},
  pages     = {29--53}
}

@book{theodoridis2024machine,
  title={Machine Learning: From the Classics to Deep Networks, Transformers, and Diffusion Models},
  author={Theodoridis, S.},
  year={2025},
  publisher={3nd Ed., Academic Press}
}

@article{genton2001classes,
  title={Classes of kernels for machine learning: a statistics perspective},
  author={Genton, Marc G},
  journal={J. Mach. Learn. Res.},
  volume={2},
  number={Dec},
  pages={299--312},
  year={2001}
}

@ARTICLE{1284395,
  author={Zhou Wang and Bovik, A.C. and Sheikh, H.R. and Simoncelli, E.P.},
  journal={IEEE Trans. Image Process.}, 
  title={Image quality assessment: from error visibility to structural similarity}, 
  year={2004},
  volume={13},
  number={4},
  pages={600-612}
}

@article{cheng2015subspace,
  title={Subspace identification for DOA estimation in massive/full-dimension MIMO systems: Bad data mitigation and automatic source enumeration},
  author={Cheng, Lei and Wu, Yik-Chung and Zhang, Jianzhong and Liu, Lingjia},
  journal={IEEE Trans. Signal Process.},
  volume={63},
  number={22},
  pages={5897--5909},
  year={2015},
  publisher={IEEE}
}

@article{tong2023bayesian,
  title={Bayesian Tensor Tucker Completion with A Flexible Core},
  author={Tong, Xueke and Cheng, Lei and Wu, Yik-Chung},
  journal={IEEE Trans. Signal Process.},
  year={2023},
  publisher={IEEE}
}

@article{xu2023tensor,
  title={Tensor train factorization under noisy and incomplete data with automatic rank estimation},
  author={Xu, Le and Cheng, Lei and Wong, Ngai and Wu, Yik-Chung},
  journal={Pattern Recognition},
  volume={141},
  pages={109650},
  year={2023},
  publisher={Elsevier}
}

@article{lee2017deep,
  title={Deep neural networks as {G}aussian processes},
  author={Lee, Jaehoon and Bahri, Yasaman and Novak, Roman and Schoenholz, Samuel S and Pennington, Jeffrey and Sohl-Dickstein, Jascha},
  journal={arXiv preprint arXiv:1711.00165},
  year={2017}
}

@article{sriperumbudur2011universality,
  title={Universality, Characteristic Kernels and RKHS Embedding of Measures.},
  author={Sriperumbudur, Bharath K and Fukumizu, Kenji and Lanckriet, Gert RG},
  journal={J. Mach. Learn. Res.},
  volume={12},
  number={7},
  year={2011}
}

@book{williams2006gaussian,
  title={Gaussian processes for machine learning},
  author={Williams, Christopher KI and Rasmussen, Carl Edward},
  year={2006},
  publisher={MIT press Cambridge, MA}
}

@inproceedings{dai2024graphical,
  title={Graphical Multioutput {G}aussian Process with Attention},
  author={Dai, Yijue and Yan, Wenzhong and Yin, Feng},
  booktitle={Int. Conf. Learn. Represent. (ICLR)},
  year={2024}
}

@article{cheng2022rethinking,
  title={Rethinking {B}ayesian learning for data analysis: The art of prior and inference in sparsity-aware modeling},
  author={Cheng, Lei and Yin, Feng and Theodoridis, Sergios and Chatzis, Sotirios and Chang, Tsung-Hui},
  journal={IEEE Signal Process. Mag.},
  volume={39},
  number={6},
  pages={18--52},
  year={2022},
  publisher={IEEE}
}

@article{li2005rank,
  title={A rank-revealing method with updating, downdating, and applications},
  author={Li, Tien-Yien and Zeng, Zhonggang},
  journal={SIAM J. Matrix Anal. Appl.},
  volume={26},
  number={4},
  pages={918--946},
  year={2005},
  publisher={SIAM}
}

@article{chandrasekaran1994rank,
  title={On rank-revealing factorisations},
  author={Chandrasekaran, Shivkumar and Ipsen, Ilse CF},
  journal={SIAM J. Matrix Anal. Appl.},
  volume={15},
  number={2},
  pages={592--622},
  year={1994},
  publisher={SIAM}
}

@article{liu2020gaussian,
  title={When {G}aussian process meets big data: A review of scalable GPs},
  author={Liu, Haitao and Ong, Yew-Soon and Shen, Xiaobo and Cai, Jianfei},
  journal={IEEE Trans. Neural Netw. Learn. Syst.},
  volume={31},
  number={11},
  pages={4405--4423},
  year={2020},
  publisher={IEEE}
}

@article{yin2020linear,
  title={Linear multiple low-rank kernel based stationary {G}aussian processes regression for time series},
  author={Yin, Feng and Pan, Lishuo and Chen, Tianshi and Theodoridis, Sergios and Luo, Zhi-Quan Tom and Zoubir, Abdelhak M},
  journal={IEEE Trans. Signal Process.},
  volume={68},
  pages={5260--5275},
  year={2020},
  publisher={IEEE}
}

@article{bro2003new,
  title={A new efficient method for determining the number of components in PARAFAC models},
  author={Bro, Rasmus and Kiers, Henk AL},
  journal={J. Chemom.},
  volume={17},
  number={5},
  pages={274--286},
  year={2003},
  publisher={Wiley Online Library}
}

@article{wax2003detection,
  title={Detection of signals by information theoretic criteria},
  author={Wax, Mati and Kailath, Thomas},
  journal={IEEE Trans. Acoust., Speech, Signal Process.},
  year={1985},
  volume={33},
  number={2},
  pages={387-392},
  publisher={IEEE}
}

@article{recht2010guaranteed,
  title={Guaranteed minimum-rank solutions of linear matrix equations via nuclear norm minimization},
  author={Recht, Benjamin and Fazel, Maryam and Parrilo, Pablo A},
  journal={SIAM Review},
  volume={52},
  number={3},
  pages={471--501},
  year={2010},
  publisher={SIAM}
}

@article{hitchcock1927expression,
  title={The expression of a tensor or a polyadic as a sum of products},
  author={Hitchcock, Frank L},
  journal={J. Math. Phys.},
  volume={6},
  number={1-4},
  pages={164--189},
  year={1927},
  publisher={Wiley Online Library}
}

@article{hinrich2020probabilistic,
  title={The probabilistic tensor decomposition toolbox},
  author={Hinrich, Jesper L and Madsen, Kristoffer H and M{\o}rup, Morten},
  journal={Machine Learning: Science and Technology},
  volume={1},
  number={2},
  pages={025011},
  year={2020},
  publisher={IOP Publishing}
}

@article{bhattacharya2015dirichlet,
  title={Dirichlet--Laplace priors for optimal shrinkage},
  author={Bhattacharya, Anirban and Pati, Debdeep and Pillai, Natesh S and Dunson, David B},
  journal={J. Am. Stat. Assoc.},
  volume={110},
  number={512},
  pages={1479--1490},
  year={2015},
  publisher={Taylor \& Francis}
}

@article{van2019shrinkage,
  title={Shrinkage priors for {B}ayesian penalized regression},
  author={Van Erp, Sara and Oberski, Daniel L and Mulder, Joris},
  journal={J. Math. Psychol.},
  volume={89},
  pages={31--50},
  year={2019},
  publisher={Elsevier}
}

@article{hoff2011separable,
  title={Separable covariance arrays via the Tucker product, with applications to multivariate relational data},
  author={Hoff, Peter D},
  journal={arXiv preprint arXiv:1008.2169},
  year={2011}
}

@article{eriksson2018scaling,
  title={Scaling {G}aussian process regression with derivatives},
  author={Eriksson, David and Dong, Kun and Lee, Eric and Bindel, David and Wilson, Andrew G},
  journal={Adv. Neural Inf. Process. Syst.},
  volume={31},
  year={2018}
}

@inproceedings{jazbec2021scalable,
  title={Scalable {G}aussian process variational autoencoders},
  author={Jazbec, Metod and Ashman, Matt and Fortuin, Vincent and Pearce, Michael and Mandt, Stephan and Ratsch, Gunnar},
  booktitle={Int. Conf. Artif. Intell. Stat.},
  pages={3511--3519},
  year={2021},
  organization={Proc. Mach. Learn. Res. (PMLR)}
}

@inproceedings{wang2020conditional,
  title={Conditional expectation propagation},
  author={Wang, Zheng and Zhe, Shandian},
  booktitle={Uncertainty in Artificial Intelligence},
  pages={28--37},
  year={2020},
  organization={Proc. Mach. Learn. Res. (PMLR)}
}

@article{cheng2022tensor,
  title={Tensor-based basis function learning for three-dimensional sound speed fields},
  author={Cheng, Lei and Ji, Xingyu and Zhao, Hangfang and Li, Jianlong and Xu, Wen},
  journal={J. Acoust. Soc. Amer.},
  volume={151},
  number={1},
  pages={269--285},
  year={2022},
  publisher={AIP Publishing}
}
\bibliographystyle{IEEEtran}

\appendices
\setcounter{section}{2}
\section{Initial Rank Selection and Running Time}

\begin{figure}[h]
	\centering
	\includegraphics[width=0.9\linewidth]{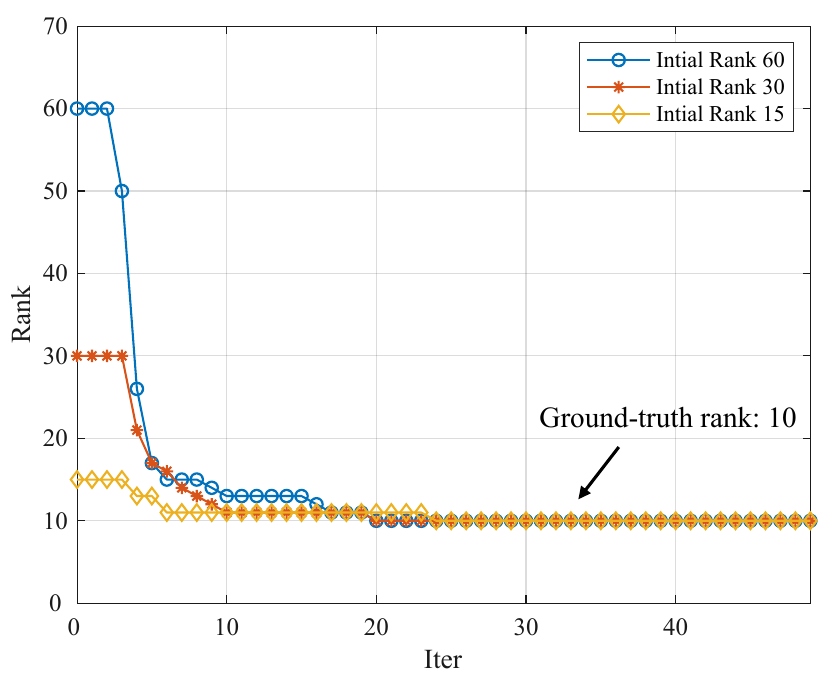}
	\caption{[Synthetic discrete data] Estimated rank versus iteration number for RR-FBTC on synthetic data. The sampling ratio is 0.3 and the SNR is 10 dB.}
	\label{fig:rankest}
\end{figure}

The proposed RR-FBTC method requires an initial rank specification. To evaluate its robustness to this initialization, we applied RR-FBTC to the synthetic discrete tensor $\boldsymbol{\mathcal{Y}} \in \mathbb{R}^{30 \times 30 \times 30}$ under various initial rank values. Fig.~\ref{fig:rankest} shows the estimated rank across iterations for three different initializations: $0.5$, $1$, and $2$ times the maximum dimension of the data tensor.

In all cases, RR-FBTC accurately estimated the true rank, demonstrating robustness to the initial choice. Additionally, the estimated rank converges rapidly to the ground truth after the initial iterations, significantly reducing computational cost. In practice, the initial rank can be set using domain knowledge if available; otherwise, a reasonable heuristic is to use the maximum dimension of the data tensor.

We compare the run times of RR-FBTC against LRTFR and FunBaT on image data, see Table~\ref{tab: rtime}. RR-FBTC achieves the fastest runtime, benefiting from both effective VI algorithm and low-rank modeling that reduces computational complexity.

\begin{table}[h]
	\centering
    \caption{Runtime of continuous tensor decomposition methods on the image data \textit{Airplane}. }
    \label{tab: rtime}
        \begin{tabular}{|c|c|c|c|}
        \hline
        Model  & LRTFR & FunBaT &  RR-FBTC \\ \hline
        Clock time & 41.7s & 194.3s &  32.6s \\ \hline
        \end{tabular}
\end{table}

\section{Ablation Study on Kernel Parameters}

In the RR-FBTC, the choice of kernel parameters plays a critical role in shaping the GP priors that govern the smoothness and characteristics of the latent functions along each mode. These parameters influence the model's ability to capture complex spatial and temporal patterns, making their selection an important consideration in practical applications. To evaluate the sensitivity to hyperparameter $h$ in (40), we conducted an ablation study using the SSF dataset under controlled conditions with a SR of 30\% and an SNR of 20 dB. We present the SSF results as a representative example, noting that multiple experiments conducted on different datasets yielded the same conclusion.

\begin{figure}[h]
	\centering
	\includegraphics[width=0.9\linewidth]{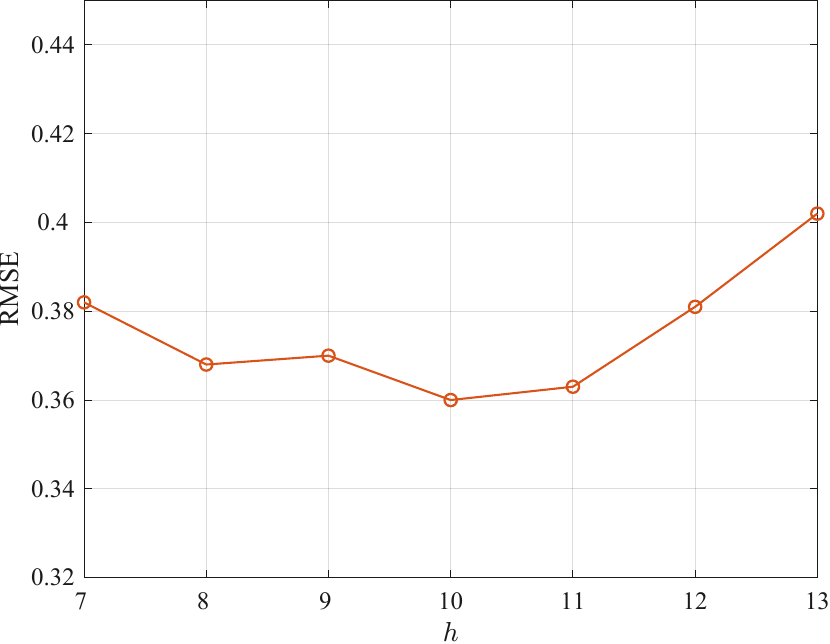}
	\caption{[SSF data] Ablation study on the kernel parameter using the SSF dataset with a SR of $30\%$ and SNR = 20 dB.}
	\label{fig:kpara}
\end{figure}

We examined the length scale parameter values to assess their impact on tensor recovery performance. 
As illustrated in Fig.~\ref{fig:kpara}, RR-FBTC exhibits notable robustness across a wide range of parameter choices, consistently achieving high reconstruction accuracy. In practice, one could further mitigate sensitivity by partitioning part of the observed data as validation data for fine-tuning the kernel parameters.

\section{Additional Results on Synthetic Data}

\begin{figure}[h]
	\centering
	\includegraphics[width=0.9\linewidth]{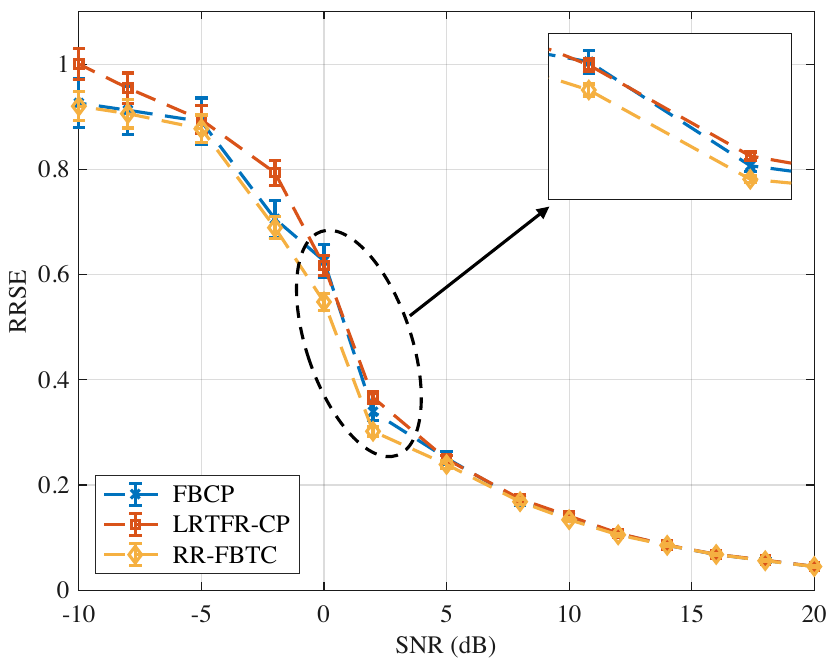}
	\caption{[Synthetic discrete data] RRSE versus SNR for different methods on the synthetic data with a SR of 0.2. LRTFR-CP uses the ground-truth rank, while FBCP and RR-FBTC automatically infer it.}
	\label{fig:rrse}
\end{figure}

To further evaluate the performace of the compared methods, we conducted extensive experiments on the synthetic tensor $\boldsymbol{\mathcal{Y}} \in \mathbb{R}^{30 \times 30 \times 30}$ under varying SNR, with a fixed SR of 0.2. The results in Fig.~\ref{fig:rrse} reveal several key insights.

All methods exhibit a characteristic sigmoid-like decrease in RRSE as SNR increases. The proposed RR-FBTC consistently achieves the lowest reconstruction error across all SNR levels This advantage stems from its ability to simultaneously estimate the tensor rank and leverage smooth functional priors, making it more resilient to noise.

Notably, when SNR falls below –10 dB, all methods fail to produce meaningful recovery, with RRSE values exceeding 1. Under such extreme noise conditions, the neural network component of LRTFR struggles to learn coherent representations, while both FBCP and RR-FBTC face challenges in accurately estimating the underlying rank.

\section{Comparison with Copula-Based Methods}

In probability theory and statistics, a copula is a powerful tool that decouples the marginal distributions of random variables from their underlying dependence structure. This flexibility allows copulas to model complex, non-Gaussian dependencies in heterogeneous datasets by combining arbitrary marginal distributions with a parametric correlation model. Recent advances have extended Gaussian copula models to low-rank matrix completion, offering a promising alternative for capturing nonlinear relationships in incomplete data.

\begin{figure}[h]
	\centering
	\includegraphics[width=0.9\linewidth]{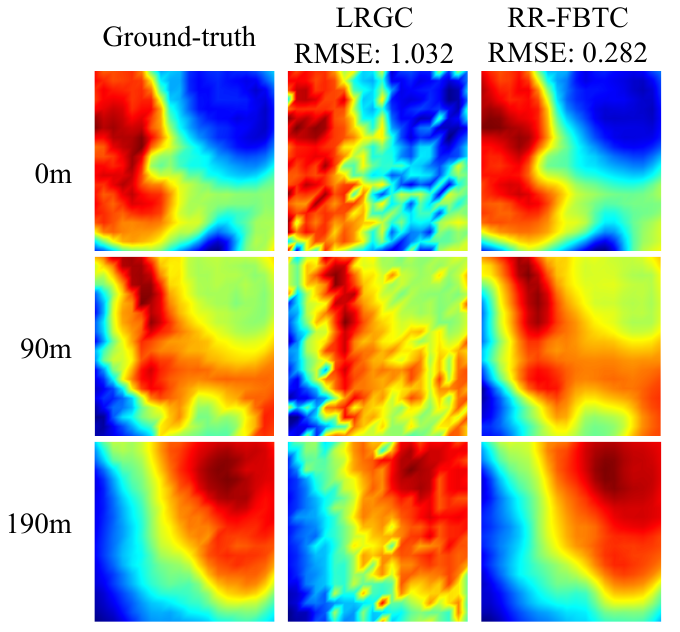}
	\caption{[SSF data] The reconstructed SSF of LRGC and RR-FBTC in one Monte Carlo trial. The SR= 50\% and SNR = 20 dB.}
	\label{fig: lrgc}
\end{figure}

Motivated by these developments, we compare our method against the Low-Rank Gaussian Copula (LRGC) approach using on-grid Sea Surface Temperature (SSF) data. The results in Table~\ref{tab:viii} indicate that RR-FBTC achieves significantly better recovery performance. We attribute this advantage to two main factors. First, while the theory guarantees the existence of a copula linking the data to a latent Gaussian variable, accurately estimating this mapping under noisy and highly incomplete observations remains challenging. Second, the LRGC model does not explicitly incorporate multi-way correlation structure across tensor modes, leading to visible artifacts in the reconstructed field, see Fig.~\ref{fig: lrgc}. In contrast, RR-FBTC leverages smoothness along continuous modes and directly models multi-dimensional dependencies, making it more suitable for tensor data with functional structure.

\begin{table}[t]
    \centering
    \caption{[SSF data] RMSEs of LRGC and RR-FBTC at different SNRs and SRs.}
    \begin{tabular}{cccc}
        \toprule
        SR & SNR & LRGC & RR-FBTC \\
        \midrule
        \multirow{2}{*}{40\%} 
        & 20 dB & 1.273$\pm$.024  & \textbf{0.316}$\pm$.013 \\
        & 10 dB & 1.602$\pm$.087  & \textbf{0.453}$\pm$.032 \\
        \midrule
        \multirow{2}{*}{50\%} 
        & 20 dB & 1.024$\pm$.022  & \textbf{0.287}$\pm$.011 \\
        & 10 dB  & 1.327$\pm$.076  & \textbf{0.329}$\pm$.038 \\
        \bottomrule
    \end{tabular}
    \label{tab:viii}
\end{table}

\end{document}